\documentclass[sigconf]{acmart}
\usepackage{amsmath,epsfig,multirow, makecell,graphicx, color,mathtools, stfloats, tabularx}

\AtBeginDocument{%
  }

\setcopyright{acmlicensed}
\copyrightyear{2024}
\acmYear{2024}
\setcopyright{acmlicensed}
\acmConference[MM '24] {Proceedings of the 32nd ACM International Conference on Multimedia}{October 28--November 1, 2024}{Melbourne, VIC, Australia.}
\acmBooktitle{Proceedings of the 32nd ACM International Conference on Multimedia (MM '24), October 28--November 1, 2024, Melbourne, VIC, Australia}
\acmISBN{979-8-4007-0686-8/24/10}

\acmDOI{10.1145/3664647.3680650}

\begin{document}

\title{Multi-Scale and Detail-Enhanced Segment Anything Model for Salient Object Detection}


\author{Shixuan Gao}
\orcid{0009-0006-4251-2495}
\affiliation{%
  \institution{Dalian University of Technology}
  \city{Dalian}
  \country{China}}
\email{gaosx@mail.dlut.edu.cn}

\author{Pingping Zhang}
\orcid{0000-0003-1206-1444}
\authornote{Corresponding author.}
\affiliation{%
  \institution{Dalian University of Technology}
  \city{Dalian}
  \country{China}}
\email{zhpp@dlut.edu.cn}

\author{Tianyu Yan}
\orcid{0009-0003-2249-3985}
\affiliation{%
  \institution{Dalian University of Technology}
  \city{Dalian}
  \country{China}}
\email{2981431354@mail.dlut.edu.cn}

\author{Huchuan Lu}
\orcid{0000-0002-6668-9758}
\affiliation{%
  \institution{Dalian University of Technology}
  \city{Dalian}
  \country{China}}
\email{lhchuan@dlut.edu.cn}

\renewcommand{\shortauthors}{Shixuan Gao, Pingping Zhang, Tianyu Yan, and Huchuan Lu.}

\begin{abstract}
Salient Object Detection (SOD) aims to identify and segment the most prominent objects in images.
Advanced SOD methods often utilize various Convolutional Neural Networks (CNN) or Transformers for deep feature extraction.
However, these methods still deliver low performance and poor generalization in complex cases.
Recently, Segment Anything Model (SAM) has been proposed as a visual fundamental model, which gives strong segmentation and generalization capabilities.
Nonetheless, SAM requires accurate prompts of target objects, which are unavailable in SOD.
Additionally, SAM lacks the utilization of multi-scale and multi-level information, as well as the incorporation of fine-grained details.
To address these shortcomings, we propose a \textbf{M}ulti-scale and \textbf{D}etail-enhanced \textbf{SAM} (MDSAM) for SOD.
Specifically, we first introduce a Lightweight Multi-Scale Adapter (LMSA), which allows SAM to learn multi-scale information with very few trainable parameters.
Then, we propose a Multi-Level Fusion Module (MLFM) to comprehensively utilize the multi-level information from the SAM's encoder.
Finally, we propose a Detail Enhancement Module (DEM) to incorporate SAM with fine-grained details.
Experimental results demonstrate the superior performance of our model on multiple SOD datasets and its strong generalization on other segmentation tasks.
The source code is released at \url{https://github.com/BellyBeauty/MDSAM}.
\end{abstract}


\maketitle

\section{Introduction}
Salient Object Detection (SOD) aims to identify and segment the most prominent objects in images.
As a fundamental task, SOD plays an important role in many downstream tasks, such as object tracking \cite{zhang2020non,zhou2021saliency}, scene segmentation \cite{wang2015saliency,zhang2019deep}, person re-identification \cite{quispe2019improved,gao2024part}, etc.
In the last decade, Convolutional Neural Networks (CNNs) have witnessed great progresses in SOD.
However, SOD requires global context information, which is challenging for CNNs due to their limited receptive fields.
Fortunately, with the global perception of self-attention, Vision Transformers (ViT)~\cite{dosovitskiy2020image} greatly prompt SOD to an expressive level.
%
%
However, due to the insufficient training examples and large domain gaps, these SOD methods still deliver low performance and poor generalization in complex cases.
\begin{figure}[t]
\centering
\includegraphics[scale=0.32]{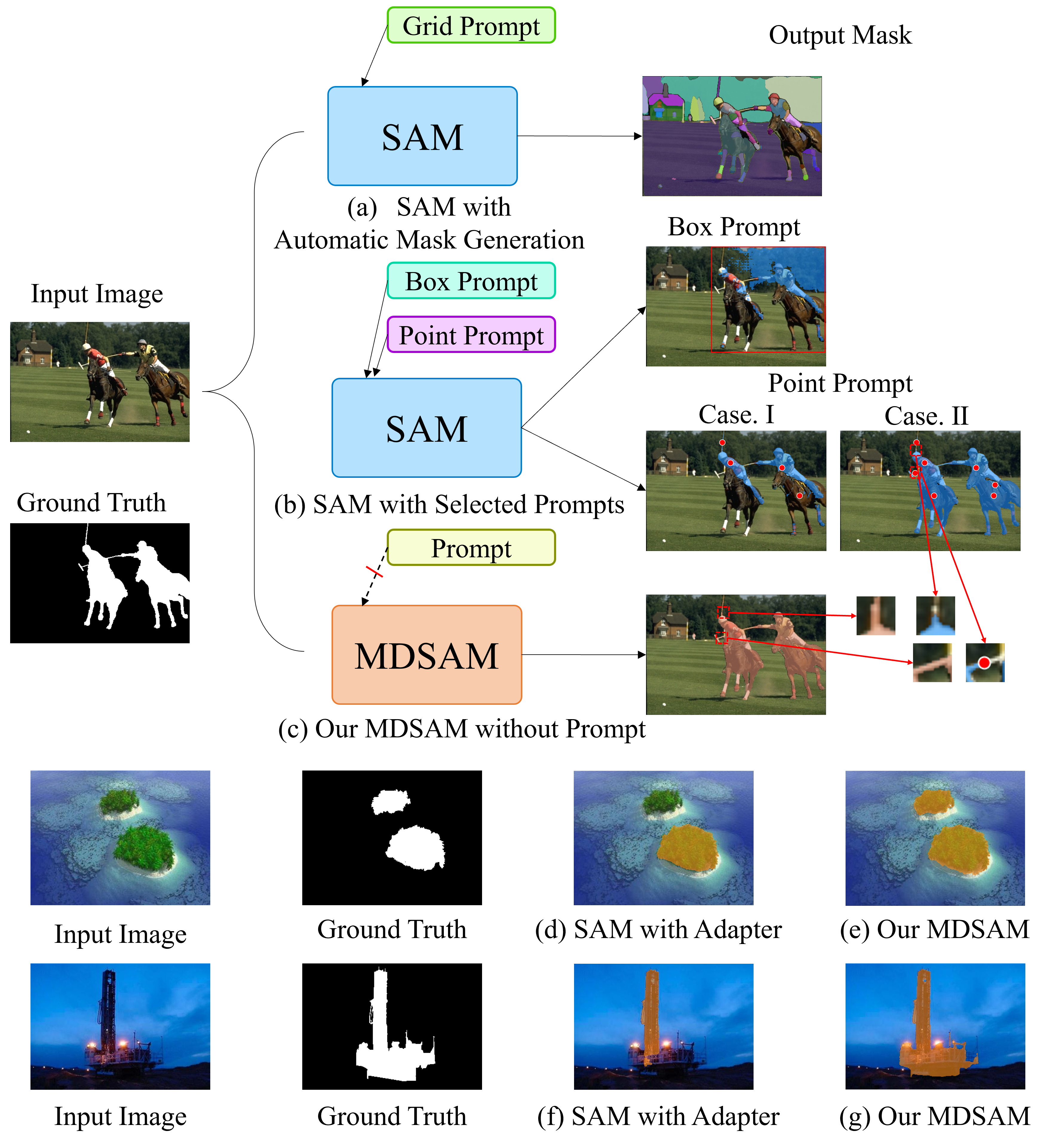}
\caption{Illustration of our motivations. The mask generation process of (a) SAM with grid prompts, (b) SAM with selected prompts (points or boxes) and (c) our MDSAM without prompts. (d) and (f) are saliency maps predicted by SAM with Adapter. (e) and (g) are saliency maps predicted by our MDSAM. Our MDSAM locate salient objects more accurately and segment them with fine-grained details.}
\label{fig:moti}
\vspace{-4mm}
\end{figure}

Recently, a visual fundamental model named Segment Anything Model (SAM)~\cite{kirillov2023segment} has been proposed for universal image segmentation.
SAM benefits from more than 1 billion training samples, which deliver a powerful generalization capability in many segmentation tasks~\cite{ma2024segment,chen2023sam,ren2024grounded,yan2024mas,zhang2024fantastic}.
%
%
%
However, to achieve robust segmentation, SAM requires handcrafted prompts such as points, boxes, or rough masks corresponding to the objects of interest.
As illustrated in Fig.~\ref{fig:moti}(a), SAM with grid prompts can automatically generate object masks.
However, these masks are class-agnostic, which are unable to identify salient objects.
As shown in Fig.~\ref{fig:moti}(b), the point prompt requires an accurate number and placement of key points.
As a result, even a slight difference can lead to incorrect results.
Meanwhile, the box prompt may be ineffective in several challenging scenes, such as object occlusion.
Thus, adapting SAM to SOD requires carefully selected prompts for salient objects.
This is improper for SOD since the ground truth is unavailable during inference.
In fact, full fine-tuning is a direct method of adapting SAM to SOD.
However, it may lead to huge training parameters, even resulting in a performance decline.
Previous works have already used Adapter~\cite{houlsby2019parameter} to transfer such fundamental models to downstream tasks.
Nonetheless, as shown in Fig.~\ref{fig:moti}(d), SAM trained with Adapters performs poorly in multi-scale scenarios.
%
%
%
Furthermore, SAM only utilizes features at the end of the image encoder, resulting in the loss of low-level information.
%
%
As illustrated in Fig.~\ref{fig:moti}(d) and (f), SAM encounters incomplete object masks and inaccurate object edges due to the lack of multi-scale information and fine-grained details.
%

To address the aforementioned issues, we propose a novel framework named Multi-scale and Detail-Enhanced SAM (MDSAM) for high-performance SOD.
Functionally, it adapts SAM to the SOD task with multi-scale and detail-enhanced information.
Specifically, we first propose a Lightweight Multi-Scale Adapter (LMSA) to adapt SAM with very few parameters while extracting multi-scale information.
Then, we propose a Multi-Level Fusion Module (MLFM) to extract and fuse the features from different levels of the SAM's encoder.
%
%
Finally, we propose a Detail Enhancement Module (DEM) to incorporate image details and edges for SOD prediction, which helps to generate precise and detailed results.
Extensive experiments demonstrate that our model not only performs well on SOD but also exhibits superior performance on other segmentation tasks.
%

Our main contributions are summarized as follows:
\begin{itemize}
\item We propose a novel framework named Multi-Scale and Detail-enhanced SAM (MDSAM) for high-performance SOD.
\item We propose a Lightweight Multi-Scale Adapter (LMSA) to adapt SAM to SOD while keeping training-efficient and strong generalization.
\item We propose Multi-Level Fusion Module (MLFM) and Detail Enhancement Module (DEM) to improve the multi-scale and fine-grained perceptions of SAM, respectively.
\item We perform extensive experiments to verify the superior effectiveness and strong generalization of our method on multiple SOD datasets and other segmentation tasks.
\end{itemize}

\begin{figure*}[t]
\centering
\includegraphics[scale=0.52]{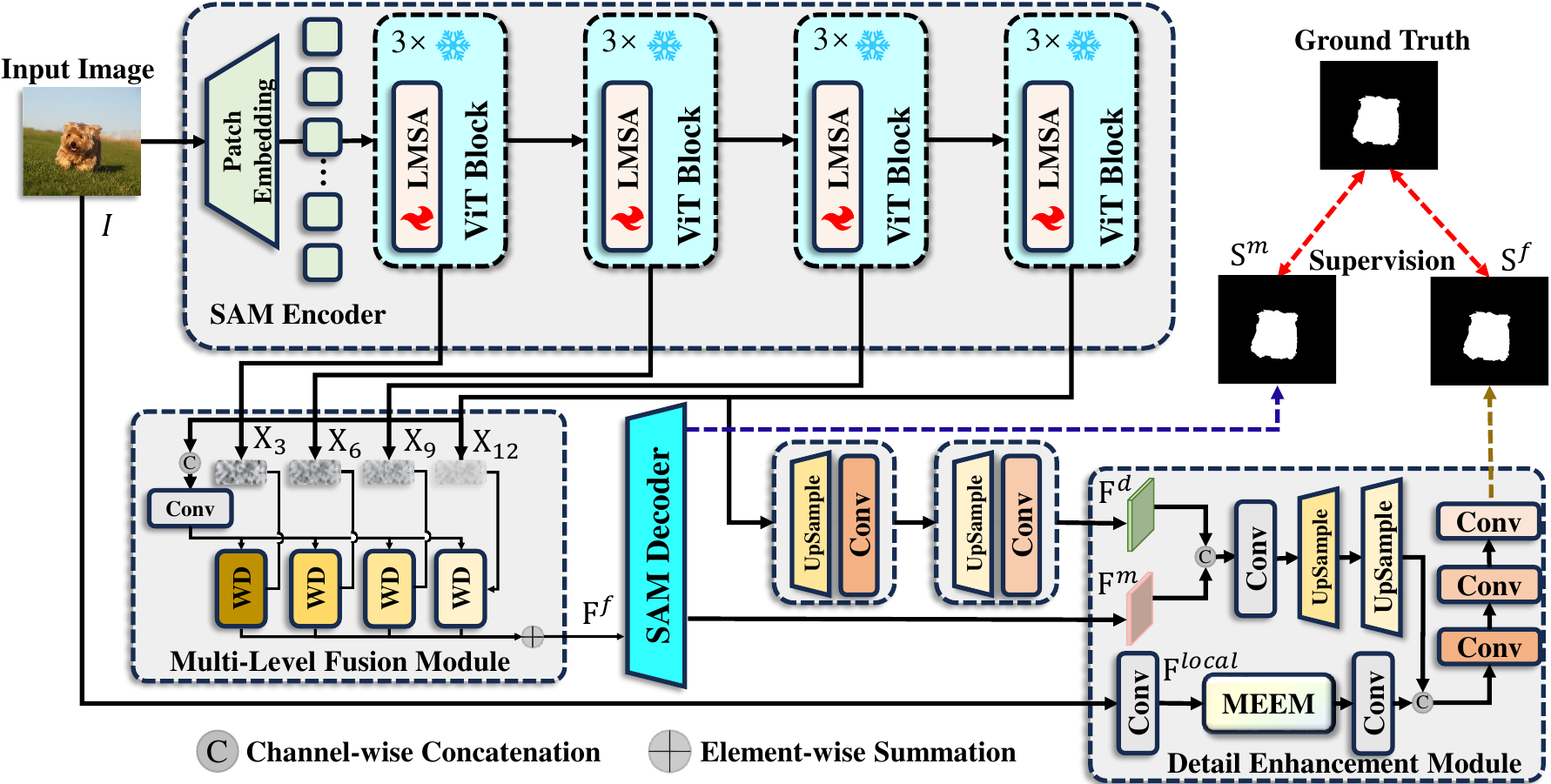}
\caption{Overall architecture of the proposed MDSAM. It reuses the pre-trained weights of SAM with three novel modules: Lightweight Multi-Scale Adapter (LMSA), Multi-Level Fusion Module (MLFM) and Detail Enhancement Module (DEM).
In addition, Weight Distributors (WD) and Multi-scale Edge Enhancement Module (MEEM) are also introduced to improve the feature representation ability.}
\vspace{-4mm}
%
\label{fig:oversll}
\end{figure*}
\section{Related Work}
\subsection{Salient Object Detection}
Currently, SOD methods are coarsely divided into two categories: CNN-based methods and Transformer-based methods.
%
%
The CNN-based methods usually adopt deep CNNs such as VGGNet~\cite{simonyan2014very}, ResNet~\cite{he2016deep}, as their backbones to extract and fuse multi-scale features.
For example, Zhang \emph{et al.}~\cite{zhang2017amulet} propose to aggregate multi-level convolutional features for SOD.
Zhang \emph{et al.}~\cite{zhang2017learning} propose to learn uncertain convolutional features for accurate SOD.
Wang \emph{et al.}~\cite{wang2017stagewise} propose a stage-wise refinement CNN model to detect salient objects in images.
Furthermore, Zhang \emph{et al.}~\cite{zhang2019salient} utilize a Siamese CNN to learn lossless feature reflection for structure-ware SOD.
Wu \emph{et al.}~\cite{wu2019cascaded} propose a cascaded partial decoder for fast and accurate SOD.
Zeng \emph{et al.}~\cite{zeng2019towards} propose a global-local CNN to fuse deep features for high-resolution SOD.
Wei \emph{et al.}~\cite{wei2020f3net} propose a fusion-feedback-focus strategy to improve the multi-level feature representation ability.
Mohammadi \emph{et al.}~\cite{mohammadi2020cagnet} introduce a content-aware guidance for SOD.
Liu \emph{et al.}~\cite{liu2020dynamic} propose a dynamic feature integration for simultaneous detection of salient objects, edges, and skeletons.
Zhao \emph{et al.}~\cite{zhao2020suppress} propose a simple gated CNN to suppress unrelated features and balance the information for SOD.
Pang \emph{et al.}~\cite{pang2020multi} introduce a multi-scale interactive network to improve the feature efficiency and prediction consistency.
Wei \emph{et al.}~\cite{wei2020label} propose to decouple the SOD labels to improve the body and detail perceptions.
Wang \emph{et al.}~\cite{wang2023pixels} propose a multi-level enhanced SOD method by integrating pixels, regions, and objects.
Although these methods achieve great progresses in SOD, they still deliver unsatisfied results.
%
%
The main reason is that CNNs essentially lack of the global perception, which is very important for SOD.

Recently, with the global perception of self-attention, ViT~\cite{dosovitskiy2020image} shows an outstanding effectiveness for the SOD task.
%
%
To take the advantages of ViT, Liu \emph{et al.}~\cite{liu2021visual} use T2T-ViT~\cite{yuan2021tokens} to capture long-range dependencies and integrate multi-level features for better SOD results.
Yun \emph{et al.}~\cite{yun2022selfreformer} propose a self-refined network with pyramid Transformers to enhance global semantic and local detail information of salient objects.
Zhuge \emph{et al.}~\cite{zhuge2022salient} utilize Swin Transformer~\cite{liu2021swin} to extract multi-scale features and enhance the integrity of detected salient regions.
Wang \emph{et al.}~\cite{wang2024learning} combine multiple Transformers to learn local-global representations for scribble-based RGB-D SOD.
Deng \emph{et al.}~\cite{deng2023recurrent} introduce a recurrent multi-scale Transformer for high-resolution SOD.
%
%
%
Despite the impressive performance, these Transformer-based methods lack of fine-grained peceptions on object details.
More importantly, they show poor generalization in complex cases.
In this work, we borrow the superior feature extraction and generalization ability of visual fundamental models, and transfer them to the SOD task for better performances.
\subsection{Segment Anything Model}
Recently, SAM~\cite{kirillov2023segment} is proposed as a visual foundation model for universal image segmentation.
With appropriate modifications, it performs remarkably well in many downstream tasks~\cite{ma2024segment,chen2023sam,ren2024grounded,yan2024mas,zhang2024fantastic}.
However, SAM requires precise prompts of the target, such as points, boxes, or masks.
These prompts are difficult to obtain for SOD.
Several works have performed the full fine-tuning on SAM in order to transfer it to the SOD task.
However, the full fine-tuning will lead to an excessive number of training parameters, even resulting in a performance decline.
Meanwhile, there are some works that attempt to transfer SAM with a small number of trainable parameters.
For example, Cui \emph{et al.}~\cite{cui2023adaptive} propose to use a Low Rank Adaptation (LoRA) of SAM to SOD.
Xu \emph{et al.}~\cite{xu2024multidimensional} introduce a multidimensional exploration of SAM for weakly-supervised SOD.
Ke \emph{et al.}~\cite{ke2024segment} design a learnable high-quality output token, which is injected into SAM's mask decoder and is responsible for predicting the high-quality mask.
%
%
%
%
%
%
Although effective, these methods fail to enable SAM to learn multi-scale and multi-level information.
%
%
%
Additionally, SAM's simple decoder fails to incorporate detailed information, resulting in inaccurate segmentation.
To address these issues, we propose LMSA to use very few training parameters to transfer SAM into SOD and enable SAM to acquire multi-scale information.
Moreover, we introduce lightweight modules to utilize fine-grained details for better SOD performance.
\section{Our Proposed Method}
In this work, we propose a novel Multi-scale and Detail-Enhanced SAM (MDSAM) for the SOD task.
Fig.~\ref{fig:oversll} shows the overall architecture, which equips SAM with three novel modules: Lightweight Multi-Scale Adapter (LMSA), Multi-Level Fusion Module (MLFM) and Detail Enhancement Module (DEM).
We will describe them in the following sections.
%
\subsection{Lightweight Multi-Scale Adapter}
Although SAM performs well in many segmentation tasks, the challenge of providing suitable prompts still limits its direct application in SOD.
One possible solution is the full fine-tuning of SAM.
However, the excessive number of trainable parameters from the SAM's encoder and insufficient SOD data may lead to unsatisfied performances.
Fortunately, Adapter~\cite{houlsby2019parameter} is an efficient method for adapting SAM to SOD with few training parameters.
In addition, the multi-scale information is very helpful for SOD.
To this end, we propose a Lightweight Multi-Scale Adapter (LMSA) to adapt SAM to SOD.
%
%
To the best of our knowledge, we are the first to apply the multi-scale adapter for transferring SAM to downstream tasks.
We make further improvements by enhancing the capability of extracting local information.
In this way, our MDSAM reuses and preserves the pre-trained weights of SAM, while incorporating multi-scale information with very few training parameters.
\begin{figure}[t]
\centering
\includegraphics[scale=0.38]{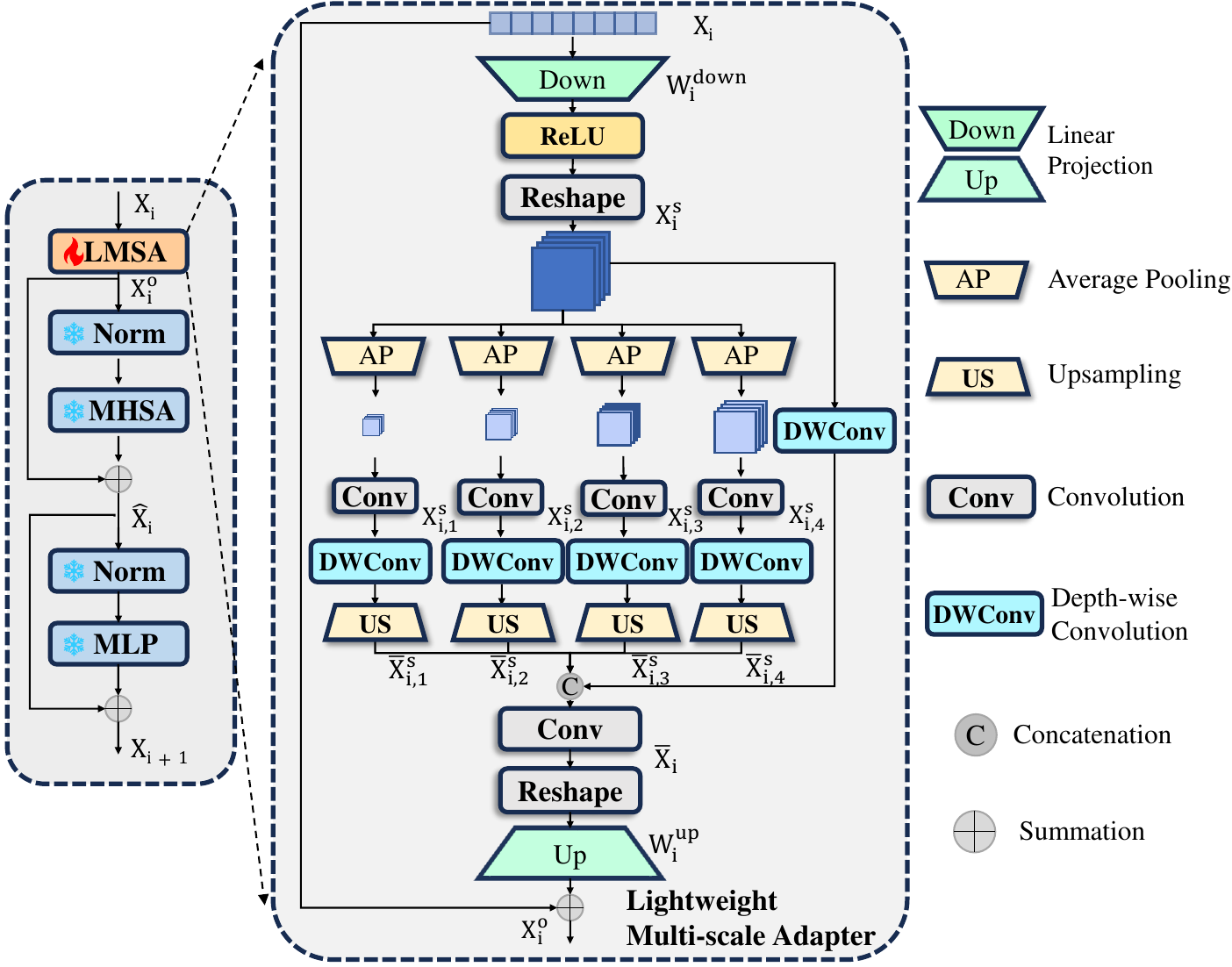}
\caption{Details of the proposed LMSA.}
\label{fig:lmsa}
\end{figure}

As shown in Fig. \ref{fig:lmsa}, each Transformer layer of the SAM's encoder is composed of a Multi-Head Self-Attention (MHSA)~\cite{vaswani2017attention}, a Multi-Layer Perceptron (MLP) and two normalization layers.
It is expressed as follows:
\begin{equation}
    \hat{\mathrm{\textbf{X}}}_i = MHSA(LN(\mathrm{\textbf{X}}_i)) + \mathrm{\textbf{X}}_i,
\end{equation}
\begin{equation}
    \mathrm{\textbf{X}}_{i + 1} = MLP(LN(\hat{\mathrm{\textbf{X}}}_i)) + \hat{\mathrm{\textbf{X}}}_i,
\end{equation}
where $\mathrm{\textbf{X}}_i \in \mathbb{R}^{N \times D} $ is the output of $i$-th Transformer layer.
$N$ is the number of tokens.
$D$ is the embedding dimension.
$\hat{\mathrm{\textbf{X}}}_i\in \mathbb{R}^{N \times D}$ is the intermediate output.
$LN$ denotes the Layer Normalization (LN)~\cite{ba2016layer}.
To adapt SAM to SOD, we employ the proposed LMSA before the first normalization in each Transformer layer.

The detailed structure of the LMSA is shown in Fig.~\ref{fig:lmsa}.
Specifically, we first use a linear projection layer to reduce the feature dimension:
\begin{equation}
    \mathrm{\textbf{X}}^{\mathrm{s}}_i = \tau(ReLU(\mathrm{\textbf{W}}^{\mathrm{down}}_i(\mathrm{\textbf{X}}_i))),
\end{equation}
where ${\mathrm{\textbf W}}^{\mathrm{down}}_i \in \mathbb{R}^{D \times \frac{D}{r}}$ is the parameter of the linear projection layer.
After a ReLU activation function, we reshape the feature into $\mathrm{\textbf{X}}^s_i \in \mathbb{R}^{\frac{D}{r} \times W \times H}$ for the further spatial information process.
$\tau[\cdot]$ is the reshape operation.
$r$ is the reduction factor.

Then, to improve the representation ability, we use four Average Pooling (AP) layers to obtain multi-scale features $\mathrm{\textbf{X}}^s_{i,j} \in \mathbb{R}^{\frac{D}{4 \times r} \times W_j \times H_j}$, and utilize depth-wise convolutional layers to capture local-detail information:
\begin{equation}
    \mathrm{\textbf{X}}^s_{i,j} = \phi_{1 \times 1}(AP(\mathrm{\textbf{X}}^s_i)), 1\le j \le 4,
\end{equation}
\begin{equation}
    \bar{\mathrm{\textbf{X}}}^s_{i,j} = US(DWConv(\mathrm{\textbf{X}}^s_{i,j})),
\end{equation}
where $\phi_{1 \times 1}$ defines a convolutional layer with $1 \times 1$ kernels followed by a GELU function~\cite{hendrycks2016gaussian}.
$DWConv$ is a depth-wise convolutional layer with $3\times3$ kernels and a GELU function.
$US$ is the bilinear interpolation for upsampling features to the specific resolution.

Afterwards, we fuse the multi-scale features $\bar{\mathrm{\textbf{X}}}^s_{i,j} \in \mathbb{R}^{\frac{D}{r} \times W \times H}$ with $\mathrm{\textbf{X}}^s_i$ as follows:
\begin{equation}
    \bar{\mathrm{\textbf{X}}}_i = \phi_{1 \times 1}([\bar{\mathrm{\textbf{X}}}^s_{i,1},\bar{\mathrm{\textbf{X}}}^s_{i,2},\bar{\mathrm{\textbf{X}}}^s_{i,3},\bar{\mathrm{\textbf{X}}}^s_{i,4},DWConv(\mathrm{\textbf{X}}^s_i)]),
\end{equation}
where $\bar{\mathrm{\textbf{X}}}_i \in \mathbb{R}^{\frac{D}{r} \times W \times H}$ and $[\cdot]$ is channel-wise concatenation.

Finally, we reshape $\bar{\mathrm{\textbf{X}}}_i$ back to the tokenized feature.
With a linear projection layer and a residual connection~\cite{he2016deep}, we obtain the final output of LMSA:
\begin{equation}
    \bar{\mathrm{\textbf{X}}}^{\mathrm{o}}_i =\mathrm{\mathbf{W}}^{\mathrm{up}}_i(\tau(\bar{\mathrm{\textbf{X}}}_i)) + \mathrm{\textbf{X}}_i,
\end{equation}
where $\mathrm{\mathbf{W}}^{\mathrm{up}}_i \in \mathbb{R}^{\frac{D}{r} \times D}$ is a linear projection layer to restore the feature dimension.

With LMSA, SAM can be adapted to the SOD task with very few training parameters.
Furthermore, compared with other methods, our method can better utilize multi-scale information, thereby enabling the model to learn better features.
\begin{figure}[t]
\centering
\includegraphics[scale=0.28]{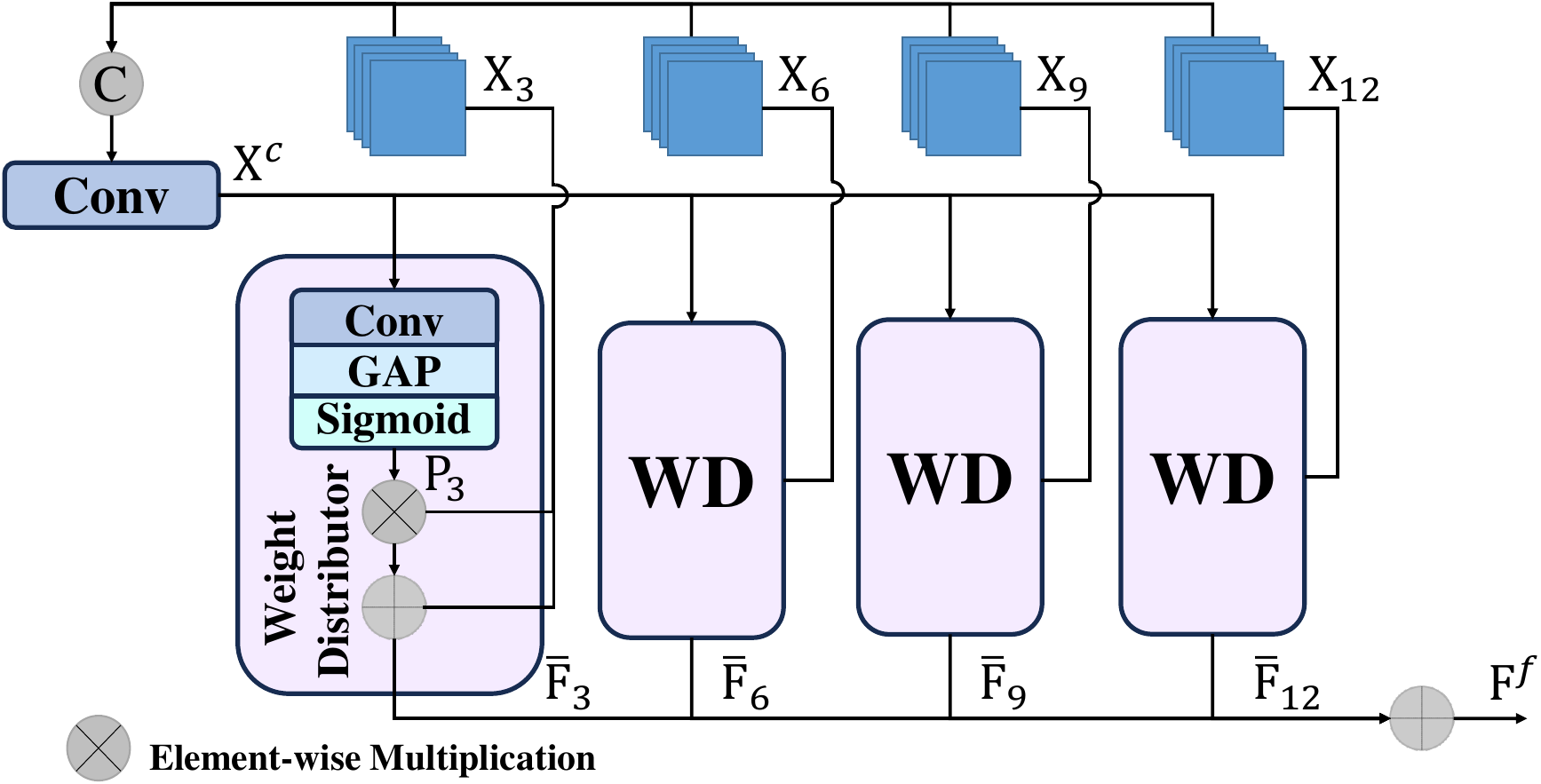}
\caption{Architecture of the proposed MLFM.}
\label{fig:mlfb}
\end{figure}
\subsection{Multi-Level Fusion Module}
In the SAM's encoder, each layer contains different information.
Shallow layers contain more low-level detail information, while deep layers contain richer high-level semantic information.
%
%
%
In the SOD task, relying solely on the high-level information from deep layers may not locate objects accurately in complex cases.
Therefore, leveraging multi-level information is necessary for SOD.
However, SAM only utilizes the output from the encoder's last layer as the mask decoder's input.
Moreover, the simple concatenation fusion strategy cannot fully integrate the multi-level information from different layers~\cite{zhang2017amulet}.
To address this issue, we propose a Multi-Level Fusion Module (MLFM) to comprehensively
utilize the multi-level information from the SAM’s encoder.
As shown in Fig.~\ref{fig:mlfb}, the proposed MLFM generates weights and assigns them to different layers with Weight Distributors (WD).
%

We denote the output features of different layers in the SAM's encoder as $\mathrm{\textbf{X}}_g \in \mathbb{R}^{D \times H \times W} (g=3,6,9,12)$.
Firstly, we concatenate them and obtain the aggregated feature $\mathrm{\mathrm{\textbf{X}}}^{c}$ by a convolutional layer:
\begin{equation}
    \mathrm{\textbf{X}}^{c} = \phi_{1 \times 1}([\mathrm{\textbf{X}}_3,\mathrm{\textbf{X}}_6,\mathrm{\textbf{X}}_9,\mathrm{\textbf{X}}_{12}]).
\end{equation}
Then, we obtain the weights $\mathrm{\textbf{P}}_g$ based on $\mathrm{\textbf{X}}^{c}$ and assign them to different layers as follows:
\begin{equation}
    \mathrm{\textbf{P}}_g = \delta(GAP(\phi_{1 \times 1}(\mathrm{\textbf{X}}^c))),
\end{equation}
\begin{equation}
    \bar{\mathrm{\textbf{F}}}_g = \mathrm{\textbf{P}}_g \times \mathrm{\textbf{X}}_g + \mathrm{\textbf{X}}_g,
\end{equation}
where $\delta$ denotes the Sigmoid function.
$GAP$ is the Global Average Pooling (GAP).
Finally, we obtain the fused feature $\mathrm{\textbf{F}}^{f} \in \mathbb{R}^{D \times H \times W}$ as follows:
\begin{equation}
    \mathrm{\textbf{F}}^{f} = \Sigma_g \bar{\mathrm{\textbf{F}}}_g.
\end{equation}
After MLFM, $\mathrm{\textbf{F}}^{f}$ will be used as the image embedding for the mask decoder.
%
%
Different from the original SAM, the output features of our proposed MLFM fully fuse multi-level information from the SAM's encoder.
\begin{figure}[t]
\centering
\includegraphics[scale=0.28]{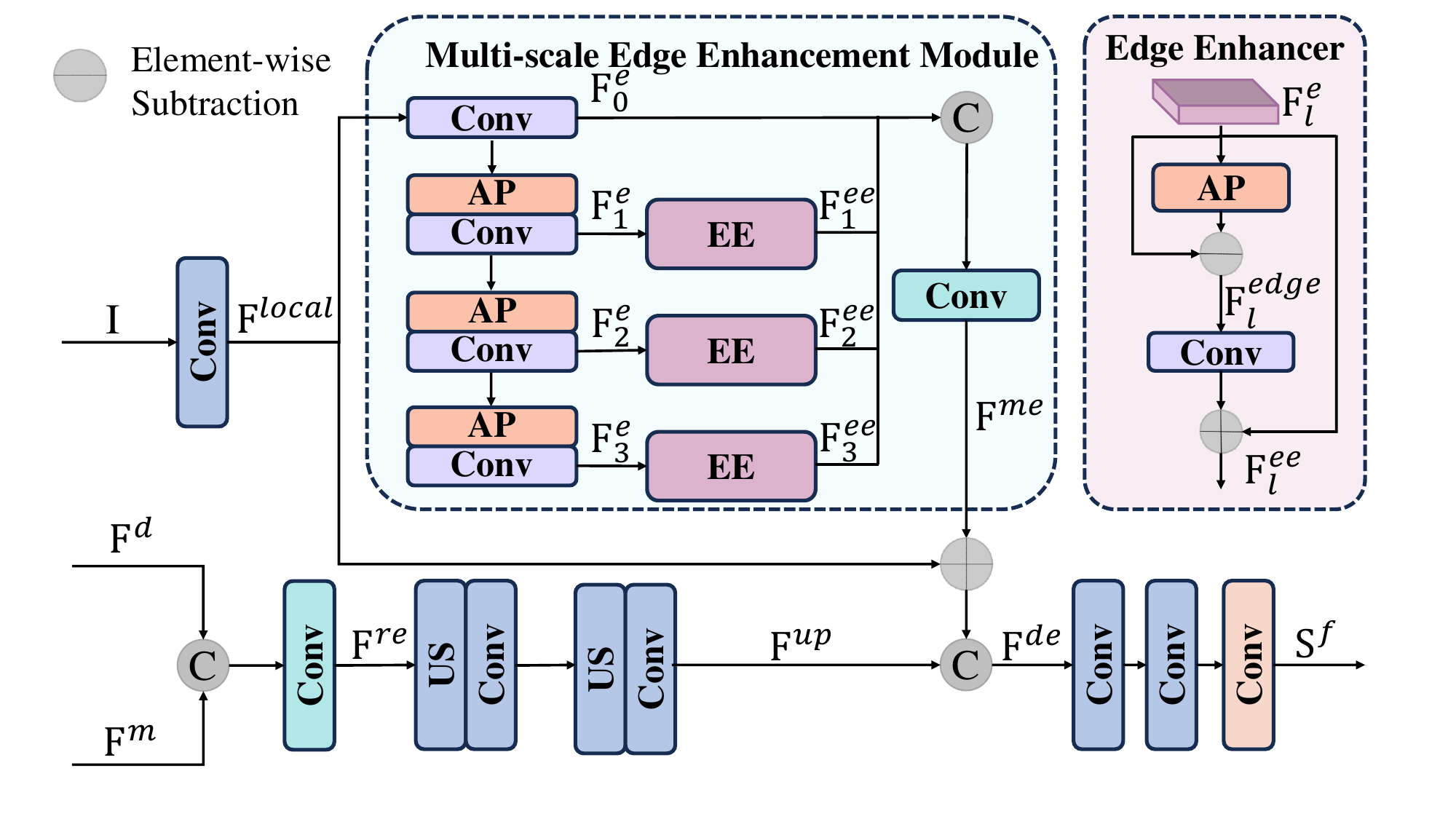}
\caption{Illustration of the proposed DEM.}
\label{fig:dem}
\end{figure}
\subsection{Detail Enhancement Module}
With the assistance of LMSA and MLFM, our framework sufficiently leverages multi-scale and multi-level information.
This greatly aids SAM's application in the SOD task.
%
%
However, there are still some remaining issues.
On the one hand, the SAM's encoder employs an image patch embedding strategy, which inevitably loses detail information.
On the other hand, the upsampling strategy in the SAM's decoder can not restore key details.
%
Thus, the salient objects with complex details and edges are not sufficiently captured.
To address this issue, we propose a Detail Enhancement Module (DEM) to enhance fine-grained details for better SOD performances.

As shown in Fig.~\ref{fig:dem}, the proposed DEM includes a primary branch and an auxiliary branch. %
The primary branch progressively upsamples features from the mask decoder's output to the input resolution.
The auxiliary branch extracts fine-grained detail information from the input image and adds it to the features in the primary branch.
However, directly extracting details at the input resolution would lead to excessive computation, slowing down the inference speed.
Therefore, we propose a Multi-scale Edge Enhancement Module (MEEM).
In MEEM, we use $3 \times 3$ average pooling and $1 \times 1$ convolutions to extract the detail information.
Additionally, we utilize Edge Enhancers (EE) to highlight object edges in the feature maps.

Technically, in the primary branch, we first concatenate the mask decoder feature $\mathrm{\textbf{F}}^{m}$ and the upsampled feature $\mathrm{\textbf{F}}^{d}$ from the last layer of SAM's encoder.
Then, we use a $1 \times 1$ convolutional layer to reduce the channel dimension.
Finally, we progressively upsample the features to the input resolution by using multiple bilinear interpolations and $3 \times 3$ convolutions:
\begin{equation}
    \mathrm{\textbf{F}}^{re} = \phi_{1 \times 1}([\mathrm{\textbf{F}}^{d}, \mathrm{\textbf{F}}^{m}]),
\end{equation}
\begin{equation}
    \mathrm{\textbf{F}}^{up} = \phi_{3 \times 3}(US_{\times 2}(\phi_{3 \times 3}(US_{\times 2}(\mathrm{\textbf{F}}^{re}))))
\end{equation}
where $US_{\times 2}$ is $2\times$ upsampling with bilinear interpolations.
$\phi_{3 \times 3}$ contains a convolutional layer with $3 \times 3$ kernels, a batch normalization, and a ReLU function.

Although $\mathrm{\textbf{F}}^{up}$ can be used for SOD prediction, it lacks of detail and edge information.
Thus, we introduce the auxiliary branch and propose a MEEM to incorporate fine-grained details from input images.
Specifically, given an input image $\mathrm{\textbf{I}}$, we first apply a $3 \times 3$ convolutional layer to extract the local features:
\begin{equation}
    \mathrm{\textbf{F}}^{local} = \phi_{3 \times 3}(\mathrm{\textbf{I}}),
\end{equation}
where $\mathrm{\textbf{F}}^{local} \in \mathbb{R}^{C \times H \times W}$.
With the proposed MEEM, we extract edge information from the image at multiple scales, and further enhance the edge perception of salient objects.
To reduce the computational complexity, we use the average pooling to expand the receptive field.
The procedure of the MEEM is as follows:
\begin{equation}
    \mathrm{\textbf{F}}^e_0 = \phi_{1 \times 1}(\mathrm{\textbf{F}}^{local}),
\end{equation}
\begin{equation}
    \mathrm{\textbf{F}}^e_{t+1} = AP(\phi_{1 \times 1}'(\mathrm{\textbf{F}}^e_{t})), (0 \le t \le 2),
\end{equation}
where $AP$ denotes the average pooling with $3 \times 3$ kernels.
$\phi_{1 \times 1}'$ denotes a $1\times1$ convolutional layer with a batch normalization and the sigmoid function.
$\mathrm{\textbf{F}}_t^e\in \mathbb{R}^{C \times H \times W}$ is the feature at scale $t$.
Then, we introduce the edge enhancer $\psi$ to strengthen the detailed information at each scale:
\begin{equation}
    \mathrm{\textbf{F}}^{ee}_{l} = \psi(\mathrm{\textbf{F}}^e_{l}), (1 \le l \le 3),
\end{equation}
where $\mathrm{\textbf{F}}_l^{ee}\in \mathbb{R}^{C \times H \times W}$ is the edge-enhanced features.
The structure of the edge enhancer is shown in the top-right part of Fig. \ref{fig:dem}, which can be represented as follows:
\begin{equation}
    \mathrm{\textbf{F}}^{\mathrm{edge}}_l = \mathrm{\textbf{F}}^e_l - AP(\mathrm{\textbf{F}}^e_l),
\end{equation}
\begin{equation}
    \mathrm{\textbf{F}}^{ee}_l = \phi_{1 \times 1}'(\mathrm{\textbf{F}}^{\mathrm{edge}}_l) + \mathrm{\textbf{F}}^e_l,
\end{equation}
where $\mathrm{\textbf{F}}_l^{\mathrm{edge}} \in \mathbb{R}^{C \times H \times W}$.
Then, we fuse these features with channel-wise concatenation and a $1 \times 1$ convolutional layer:
\begin{equation}
    \mathrm{\textbf{F}}^{me} = \phi_{1 \times 1}([\mathrm{\textbf{F}}^e_0, \mathrm{\textbf{F}}^{ee}_1, \mathrm{\textbf{F}}^{ee}_2, \mathrm{\textbf{F}}^{ee}_3]),
\end{equation}
where $\mathrm{\textbf{F}}^{me}\in \mathbb{R}^{C \times H \times W}$ is the feature output of the MEEM.
In this way, $\mathrm{\textbf{F}}^{me}$ includes both the fine-grained details and the multi-scale edge information.
We use these features to complement the missing information in feature $\mathrm{\textbf{F}}^{c}$.
After concatenation, we apply two $3\times3$ convolutional layers and a $1\times1$ convolutional layer to obtain the final SOD result $\mathrm{\textbf{S}}^f$:
\begin{equation}
    \mathrm{\textbf{F}}^{de} = [\mathrm{\textbf{F}}^{up},
    \mathrm{\textbf{F}}^{me} + \mathrm{\textbf{F}}^{local}],
\end{equation}
\begin{equation}
    \mathrm{\textbf{S}}^f = \phi_{1 \times 1}(\phi_{3 \times 3}(\phi_{3 \times 3}(\mathrm{\textbf{F}}^{de}))).
\end{equation}

As can be observed, with the assistance of MEEM, our DEM can extract multi-scale edges from input images and combine them with the primary branch.
By utilizing the two branches, the issue of lacking details in SAM has been resolved.
As a result, our MDSAM can effectively locate salient objects with rich detail information.
\subsection{Loss Functions}
To train our framework, we introduce the Binary Cross Entropy (BCE) loss, the Interaction over Union (IoU) loss, and the L1 loss.
%
To improve the learning ability, we employ them to both $\mathrm{\textbf{S}}^f$ and $\mathrm{\textbf{S}}^m=\phi_{1 \times 1}(\mathrm{\textbf{F}}^m)$.
The total loss of MDSAM is formulated as follows:
\begin{equation}
    \mathcal{L}(\mathrm{\textbf{S}},\mathrm{\textbf{S}}_{gt})=\mathcal{L}_{BCE} + \mathcal{L}_{IoU} + \mathcal{L}_{L1},
\end{equation}
\begin{equation}
    \mathcal{L}_{total} = \mathcal{L}_f(\mathrm{\textbf{S}}^f,\mathrm{\textbf{S}}^{gt}) + \mathcal{L}_m(\mathrm{\textbf{S}}^m,\mathrm{\textbf{S}}^{gt}),
\end{equation}
where $\mathrm{\textbf{S}}^{gt}$ is the ground truth of salient objects.
\begin{table*}[t]
\begin{center}
\caption{Quantitative comparison between our method and other methods. The best, second, and third results are highlighted in {\color{red}{red}}, {\color{green}{green}}, and {\color{blue}{blue}}, respectively.}
\label{tab:comp1}
\resizebox{0.98\textwidth}{!}{
    \begin{tabular}{c|cccc|cccc|cccc}
      \hline
      \multirow{2}{*}{\textbf{Method}}
      & \multicolumn{1}{c}{\multirow{2}{*}{\textbf{Input Size}}}
      & \multicolumn{1}{c}{\multirow{2}{*}{\textbf{Params (M)}}}
      & \multicolumn{1}{c}{\multirow{2}{*}{\textbf{FLOPs (G)}}}
      & \multicolumn{1}{c|}{\multirow{2}{*}{\textbf{FPS}}}
      & \multicolumn{4}{c|}{\textbf{DUTS-TE}} & \multicolumn{4}{c}{\textbf{DUT-OMRON}}\\

      & & & & & $MAE$ & $F^{max}_\beta$ & $S_m$ & $E_m$ &
      $MAE$ & $F^{max}_\beta$ & $S_m$ & $E_m$  \\
      \hline

      \multicolumn{13}{c}{CNN-Based Methods} \\
      \hline

      CPD$_{2019}$~\cite{wu2019cascaded} & $352 \times 352$ & 47.84 & 17.82 & 123
      & 0.043 & 0.972 & 0.869 & 0.898
      & 0.056 & 0.818 & 0.825 & 0.847\\

      F3Net$_{2020}$~\cite{wei2020f3net} & $352 \times 352$ & 25.54 & 16.48 & 167
      & 0.035 & 0.905 & 0.888 & 0.920
      & 0.053 & 0.841 & 0.838 & 0.864\\

      CAGNet-L$_{2020}$~\cite{mohammadi2020cagnet} & $480 \times 480$ & - & - & -
      & 0.029 & 0.898 & 0.897 & 0.939
      & 0.047 & 0.818 & 0.845 & 0.882\\

      DFI$_{2020}$~\cite{liu2020dynamic} & $224 \times 224$ & 29.61 & 11.31 & 102
      & 0.039 & 0.896 & 0.887 & 0.912
      & 0.055 & 0.818 & 0.839 & 0.865\\

      GateNet-X$_{2020}$~\cite{zhao2020suppress} & $384 \times 384$ & 128.63 & 162.13 & 130
      & 0.035 & 0.908 & 0.897 & 0.916
      & 0.051 & 0.847 & 0.849 & 0.865\\

      MINet-R$_{2020}$~\cite{pang2020multi} & $320 \times 320$ & 162.38 & 87.10 & 62
      & 0.037 & 0.884 & 0.884 & 0.917
      & 0.056 & 0.831 & 0.833 & 0.860\\

      LDF$_{2020}$~\cite{wei2020label} & $352 \times 352$ & 25.15 & 15.57 & 177
      & 0.034 & 0.905 & 0.892 & 0.925
      & 0.052 & 0.835 & 0.839 & 0.865\\

      TE3$_{2022}$~\cite{lee2022tracer} & $384 \times 384$ & 14.02 & 3.23 & 24
      & 0.028 & 0.909 & 0.899 & 0.943
      & 0.046 & 0.840 & 0.848 & 0.881 \\

      TE5$_{2022}$~\cite{lee2022tracer} & $512 \times 512$ & 31.30 & 6.06 & 15
      & 0.026 & 0.923 & 0.910 & 0.948
      & 0.045 & 0.850 & 0.856 & 0.887\\

      TE7$_{2022}$~\cite{lee2022tracer} & $640 \times 640$ & 66.27 & 10.17 & 9
      & \color{red}0.023 & \color{blue}0.932 & \color{green}0.920 & \color{red}0.954
      & 0.045 & 0.849 & 0.856 & 0.883\\

      MENet$_{2023}$~\cite{wang2023pixels} & $354 \times 354$ & - & - & -
      & 0.028 & 0.918 & 0.905 & 0.938
      & 0.045 & 0.845 & 0.850 & 0.871\\

      \hline
      \multicolumn{13}{c}{Transformer-Based Methods} \\
      \hline

      VST$_{2021}$~\cite{liu2021visual} & $224 \times 224$ & 44.48 & 23.18 & 70
      & 0.037 & 0.895 & 0.896 & 0.919
      & 0.058 & 0.836 & 0.850 & 0.871\\

      SelfReformer$_{2022}$~\cite{yun2022selfreformer} & $224 \times 224$ & 90.70 & 12.83 & 62
      & 0.027 & 0.920 & 0.911 & 0.943
      & \color{blue}0.043 & 0.853 & 0.861 & 0.884\\

      ICON-S$_{2022}$~\cite{zhuge2022salient} & $384 \times 384$ & 92.15 & 52.80 & 69
      & \color{blue}0.025 & 0.924 & 0.917 & \color{red}0.954
      & \color{blue}0.043 & 0.862 & 0.869 & \color{blue}0.900\\

      BBRF$_{2023}$~\cite{ma2023boosting} & $352 \times 352$ & 74.00 & 67.02 & 62
      & \color{blue}0.025 & 0.911 & 0.909 & 0.949
      & 0.044 & 0.839 & 0.861 & 0.896\\

      DC-Net-S$_{2023}$~\cite{zhu2023dc} & $384 \times 384$ & 509.61 & 211.27 & 24
      & \color{red}0.023 & \color{blue}0.932 & \color{red}0.925 & \color{green}0.952
      & \color{red}0.039 & \color{blue}0.868 & \color{blue}0.875 & 0.898\\

      SAM$_{2023}$~\cite{kirillov2023segment} & $512 \times 512$ & 89.94 & 103.17 & 46
      & 0.030 & 0.921 & 0.909 & 0.937
      & 0.044 & 0.865 & 0.869 & 0.896\\

      \hline
      \textbf{MDSAM}$_{2024}$ & $384 \times 384$ & 100.21 & 66.23 & 50
      & \color{blue}0.025 & \color{green}0.934 & \color{blue}0.919 & \color{blue}0.950
      & \color{green}0.040 & \color{green}0.886 & \color{red}0.881 & \color{red}0.913\\

      \textbf{MDSAM}$_{2024}$ & $512 \times 512$ & 100.21 & 123.44 & 35
      & \color{green}0.024 & \color{red}0.937 & \color{green}0.920 & 0.949
      & \color{red}0.039 & \color{red}0.887 & \color{green}0.878 & \color{green}0.910\\
      \hline
    \end{tabular}
}
\end{center}
\end{table*}
\begin{table*}[t]
\begin{center}
\caption{Quantitative comparison between our method and other SOTA methods. The best, second, and third results are highlighted in {\color{red}{red}}, {\color{green}{green}}, and {\color{blue}{blue}}, respectively.}
\label{tab:comp2}
\resizebox{0.9\textwidth}{!}{
    \begin{tabular}{c|cccc|cccc|cccc}
      \hline
      \multirow{2}{*}{\textbf{Method}}
      &  \multicolumn{4}{c|}{\textbf{HKU-IS}}& \multicolumn{4}{c|}{\textbf{ECSSD}}& \multicolumn{4}{c}{\textbf{PASCAL-S}}\\

      &$MAE$ & $F^{max}_\beta$ & $S_m$ & $E_m$ &
      $MAE$ & $F^{max}_\beta$ & $S_m$ & $E_m$ &
      $MAE$ & $F^{max}_\beta$ & $S_m$ & $E_m$\\
      \hline

      \multicolumn{13}{c}{CNN-Based Methods} \\
      \hline

      CPD$_{2019}$~\cite{wu2019cascaded}
      & 0.034 & 0.828 & 0.905 & 0.938
      & 0.037 & 0.946 & 0.918 & 0.942
      & 0.071 & 0.876 & 0.848 & 0.882\\

      F3Net$_{2020}$~\cite{wei2020f3net}
      & 0.028 & 0.943 & 0.917 & 0.952
      & 0.033 & 0.957 & 0.924 & 0.948
      & 0.061 & 0.892 & 0.861 & 0.898\\

      CAGNet-L$_{2020}$~\cite{mohammadi2020cagnet}
      & 0.024 & 0.940 & 0.923 & 0.961
      & 0.026 & 0.950 & 0.930 & 0.959
      & 0.063 & 0.878 & 0.870 & 0.917\\

      DFI$_{2020}$~\cite{liu2020dynamic}
      & 0.031 & 0.934 & 0.920 & 0.951
      & 0.035 & 0.949 & 0.927 & 0.924
      & 0.065 & 0.885 & 0.857 & 0.861\\

      GateNet-X$_{2020}$~\cite{zhao2020suppress}
      & 0.029 & 0.946 & 0.925 & 0.947
      & 0.035 & 0.957 & 0.929 & 0.944
      & 0.064 & 0.892 & 0.865 & 0.895\\

      MINet-R$_{2020}$~\cite{pang2020multi}
      & 0.029 & 0.942 & 0.919 & 0.952
      & 0.033 & 0.954 & 0.925 & 0.950
      & 0.064 & 0.881 & 0.856 & 0.896\\

      LDF$_{2020}$~\cite{wei2020label}
      & 0.028 & 0.943 & 0.919 & 0.953
      & 0.034 & 0.956 & 0.924 & 0.948
      & 0.060 & 0.887 & 0.863 & 0.903\\

      TE3$_{2022}$~\cite{lee2022tracer}
      & 0.025 & 0.944 & 0.924 & 0.961
      & 0.029 & 0.954 & 0.929 & 0.958
      & 0.052 & 0.896 & 0.871 & 0.916 \\

      TE5$_{2022}$~\cite{lee2022tracer}
      & 0.022 & 0.950 & 0.930 & 0.963
      & 0.027 & 0.959 & 0.934 & 0.958
      & 0.050 & 0.900 & 0.879 & 0.921\\

      TE7$_{2022}$~\cite{lee2022tracer}
      & \color{blue}0.021 & 0.953 & \color{blue}0.934 & 0.967
      & 0.026 & 0.962 & 0.936 & 0.959
      & \color{red}0.047 & 0.906 & \color{blue}0.883 & \color{red}0.928\\

      MENet$_{2023}$~\cite{wang2023pixels}
      & 0.023 & 0.951 & 0.927 & 0.960
      & \color{red}0.021 & 0.957 & 0.928 & 0.951
      & 0.053 & 0.897 & 0.872 & 0.910\\

      \hline
      \multicolumn{13}{c}{Transformer-Based Methods} \\
      \hline

      VST$_{2021}$~\cite{liu2021visual}
      & 0.029 & 0.946 & 0.928 & 0.952
      & 0.033 & 0.954 & 0.932 & 0.951
      & 0.061 & 0.882 & 0.872 & 0.902\\

      SelfReformer$_{2022}$~\cite{yun2022selfreformer}
      & 0.024 & 0.949 & 0.931 & 0.960
      & 0.027 & 0.959 & 0.936 & 0.957
      & 0.051 & 0.902 & 0.881 & 0.919\\

      ICON-S$_{2022}$~\cite{zhuge2022salient}
      & 0.022 & 0.954 & \color{green}0.935 & \color{blue}0.968
      & \color{blue}0.023 & 0.962 & 0.914 & \color{green}0.968
      & \color{green}0.048 & 0.903 & \color{green}0.885 & \color{green}0.924\\

      BBRF$_{2023}$~\cite{ma2023boosting}
      & \color{green}0.020 & 0.949 & 0.932 & \color{green}0.969
      & \color{green}0.022 & 0.961 & 0.939 & \color{red}0.969
      & \color{blue}0.049 & 0.887 & 0.878 & \color{blue}0.923\\

      DC-Net-S$_{2023}$~\cite{zhu2023dc}
      & \color{blue}0.021 & \color{blue}0.957 & \color{red}0.941 & 0.966
      & \color{blue}0.023 & \color{blue}0.968 & \color{green}0.947 & 0.965
      & \color{blue}0.049 & \color{blue}0.904 & \color{red}0.887 & 0.917\\

      SAM$_{2023}$~\cite{kirillov2023segment}
      & 0.022 & 0.956 & \color{green}0.935 & 0.965
      & 0.025 & \color{blue}0.968 & 0.944 & 0.964
      & 0.061 & 0.897 & 0.866 & 0.902\\

      \hline
      \textbf{MDSAM}$_{2024}$
      & \color{green}0.020 & \color{green}0.962 & \color{red}0.941 & \color{red}0.970
      & \color{blue}0.023 & \color{green}0.972 & \color{blue}0.946 & 0.965
      & 0.052 & \color{red}0.912 & 0.880 & 0.918\\

      \textbf{MDSAM}$_{2024}$
      & \color{red}0.019 & \color{red}0.963 & \color{red}0.941 & \color{green}0.969
      & \color{red}0.021 & \color{red}0.974 & \color{red}0.948 & \color{blue}0.967
      & 0.052 & \color{green}0.907 & 0.882 & 0.917\\
      \hline
    \end{tabular}
}
\end{center}
\end{table*}
\begin{figure*}[t]
\centering
\includegraphics[scale=0.62]{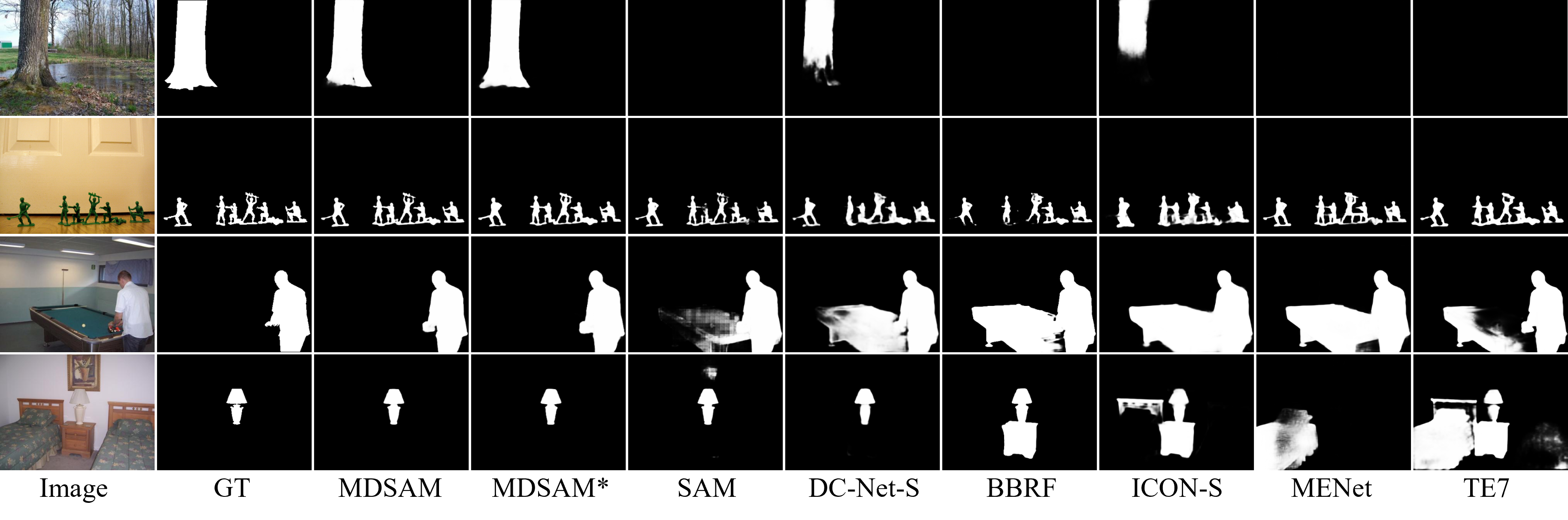}
\vspace{-4mm}
\caption{Visual comparison of saliency maps with our MDSAM and other methods. \textbf{MDSAM} is with a $512 \times 512$ input resolution. \textbf{MDSAM*} is with a $384 \times 384$ input resolution.}
\label{fig:qual}
\end{figure*}
\section{Experiments}
\subsection{Experiment Settings}
\textbf{Datasets.}
For fair comparisons, we train our proposed MDSAM on DUTS-TR~\cite{wang2017learning} (10533 images), and evaluate it on five SOD benchmark datasets, including DUTS-TE~\cite{wang2017learning} (5019 images), DUTS-OMRON~\cite{yang2013saliency} (5168 images), HKU-IS~\cite{li2015visual} (4447 images), ECSSD~\cite{yan2013hierarchical} (1000 images) and PASCAL-S~\cite{li2014secrets} (850 images).

\textbf{Metrics.}
Following previous works, we adopt four widely-used metrics to evaluate the SOD performance, i.e., the Mean Absolute Error ($MAE$)~\cite{perazzi2012saliency}, the max F-measure ($F^{max}_\beta$)~\cite{yang2013saliency}, the S-measure ($S_m$)~\cite{fan2017structure} and the mean Enhanced-alignment Measure ($E_m$)~\cite{fan2018enhanced}.

\textbf{Implementation Details.}
The PyTorch toolbox is used to implement our method with an NVIDIA A100 GPU.
For initialization, we load the weights of the image encoder and mask decoder from the SAM-B model.
And the rest of our proposed MDSAM is initialized randomly.
We resize the images into $512\times512$ and $384\times384$ as the input and set the batch sizes to 16 and 32, respectively.
We train the model using the AdamW optimizer with a weight decay of $1e^{-4}$.
During training, we freeze SAM’s encoder and set the learning rate to $5e^{-5}$ for the rest of the pre-trained weights.
For our proposed modules, we set the learning rate to $5e^{-4}$.
We employ a warm-up period of 5 epochs and train until the maximum of 80 epochs.
\subsection{Comparison with the State-of-the-arts}
We compare our proposed MDSAM with 15 other models, including CPD~\cite{wu2019cascaded}, F3Net~\cite{wei2020f3net}, CAGNet~\cite{mohammadi2020cagnet}, DFI~\cite{liu2020dynamic}, GateNet~\cite{zhao2020suppress}, MINet~\cite{pang2020multi}, LDF~\cite{wei2020label}, ICON~\cite{zhuge2022salient}, TE~\cite{lee2022tracer}, MENet~\cite{wang2023pixels}, VST~\cite{liu2021visual}, SelfReformer~\cite{yun2022selfreformer}, DC-Net~\cite{zhu2023dc}, BRRF~\cite{ma2023boosting}, SAM~\cite{kirillov2023segment}.
Note that, the SAM keeps the original structure and is fully fine-tuned for SOD.
For a fair comparison, the saliency maps are either provided by authors or generated by their released pre-trained model.
All metrics are calculated by the same tool.

\textbf{Quantitative Evaluation.}
Tab. \ref{tab:comp1} and Tab. \ref{tab:comp2} show the quantitative results of the compared methods.
Our proposed MDSAM with $512 \times 512$ input resolution achieves the best results on DUTS-OMRON, HKU-IS and ECSSD.
Furthermore, our MDSAM demonstrates a high competitiveness on the DUTS as well.
Although our MDSAM delivers inferior performance on the PASCAL-S dataset, it achieves the best overall results.
With the $384 \times 384$ input resolution, our MDSAM attains the best overall performance around similar resolutions.
In Tab.~\ref{tab:comp1} and Tab.~\ref{tab:comp2}, SAM is fully fine-tuned at a resolution of $512 \times 512$.
It can be observed that compared with the original SAM, our MDSAM significantly improves the performance while slightly increasing the model parameters.
At the same resolution, the inference speed of our MDSAM only experiences a slight decrease.
Moreover, when our MDSAM infers at a resolution of $384 \times 384$, it outperforms SAM at a resolution of $512 \times 512$ in both inference speed and accuracy.

\textbf{Qualitative Evaluation}.
Fig.~\ref{fig:qual} illustrates the predicted saliency maps with our MDSAM and other methods.
In complex scenarios, our MDSAM can accurately locate salient objects of various sizes and fully recognize the shape of these objects.
Furthermore, the results of our proposed MDSAM display more fine-grained details and accurate edges than other methods.
More visual examples can be found in Appendixes.
\subsection{Ablation Studies}
To verify the effectiveness of our proposed modules, we conduct ablation studies in this section.
If not specified, all the results are obtained with an image resolution of $512 \times 512$.

\begin{figure}[t]
\centering
\includegraphics[scale=0.42]{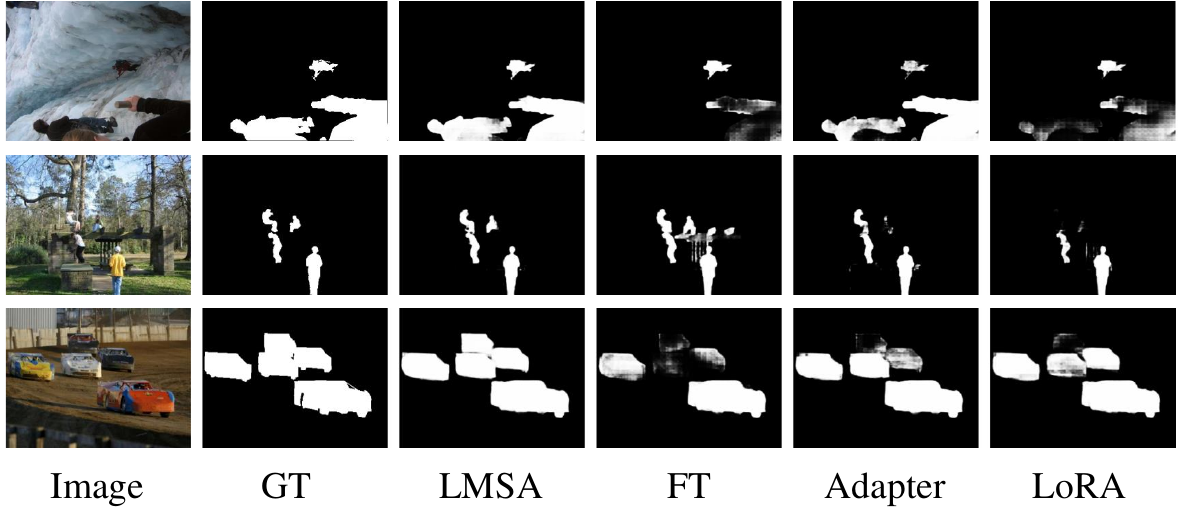}
\caption{Visual comparison of different adapting strategies for SAM on SOD.}
\label{fig:ab1}
\end{figure}

\textbf{Effectiveness of LMSA.}
To verify the effect of LMSA, we only change the image encoder of SAM.
Except of full fine-tuning (FT), we introduce Adapter~\cite{houlsby2019parameter} and LoRA~\cite{hu2021lora} to the image encoder for adapting SAM to SOD.
In addition, we keep the parameters of the Adapter, LoRA similar to our LMSA for a fair comparison.
As shown in Tab.~\ref{tab:ab1}, the introduction of parameter efficient fine-tuning can significantly reduce the trainable parameters and improve the performance.
Furthermore, our LMSA has similar parameters to the Adapter and LoRA.
However, it shows better results than them.
%
The visual comparison in Fig. \ref{fig:ab1} further illustrates that the utilization of our LMSA enables the model to acquire multi-scale information.
Consequently, it can accurately locate salient objects of varying sizes and quantities in complex scenarios.
\begin{figure*}[t]
\centering
\includegraphics[width=0.9\linewidth]{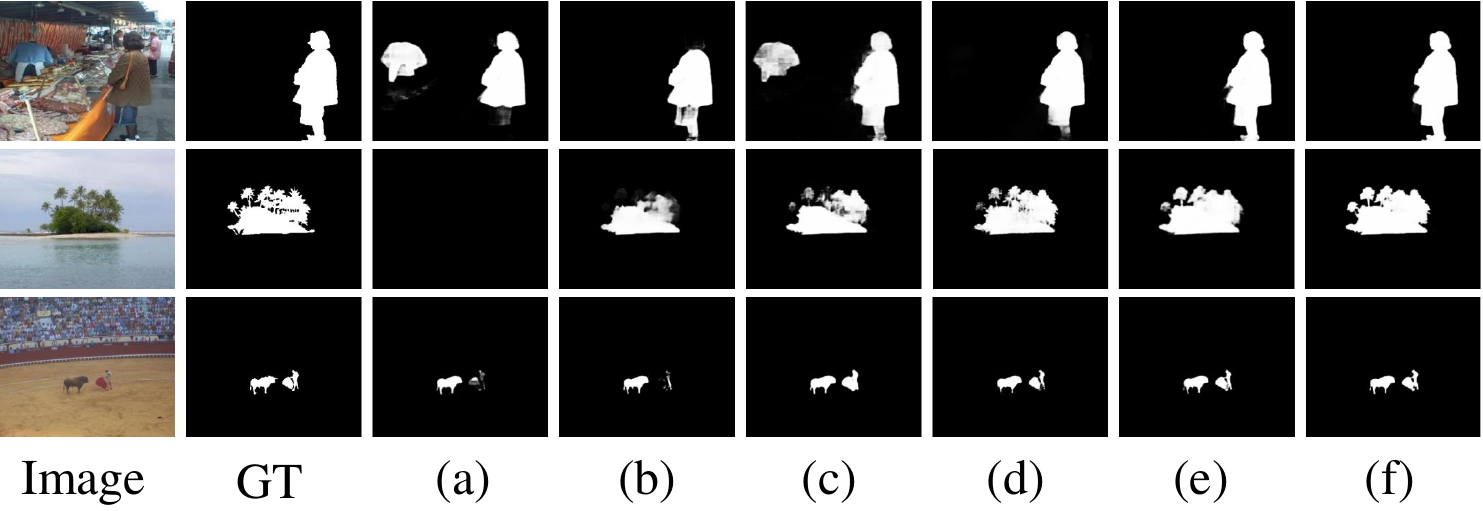}
\caption{Visual comparison of our proposed modules. The corresponding models are listed in Tab.~\ref{tab:ab2}. }
\label{fig:ab2}
\end{figure*}
\begin{table}[t]
\renewcommand{\arraystretch}{1.2}
\setlength{\tabcolsep}{1pt}
\centering
\caption{Ablation studies of the LMSA. * denotes the parameters of SAM's mask decoder.}
\label{tab:ab1}
\resizebox{0.46\textwidth}{!}{
\begin{tabular}{c|c|ccc|ccc}
     \hline
     \multirow{2}{*}{\textbf{Method}} & \textbf{Trainable}
     & \multicolumn{3}{c|}{\textbf{DUTS-TE}} & \multicolumn{3}{c}{\textbf{DUT-OMRON}} \\
     & \textbf{Parameters (M)} & $MAE$ & $F^{max}_\beta$ & $S_m$ & $MAE$ & $F^{max}_\beta$ & $S_m$ \\
     \hline
     Full fine-tuning &83.43 + 3.51*& 0.030 & 0.921 & 0.909 & 0.044 & 0.865 & 0.869 \\

     Adapter~\cite{houlsby2019parameter} & 7.09 + 3.51* & 0.028 & 0.923 & 0.915 & 0.045 & 0.866 & 0.871 \\

     LoRA~\cite{hu2021lora}& 7.09 + 3.51* &0.028 & 0.924 & 0.914 & 0.044 & 0.864 & 0.872 \\
     \hline
     LMSA & 7.15 + 3.51*  & \textbf{0.027} & \textbf{0.927} & \textbf{0.917} & \textbf{0.043} & \textbf{0.872} & \textbf{0.874} \\
     \hline
\end{tabular}
}
\end{table}

\noindent\textbf{Effectiveness of MLFM.}
Tab. \ref{tab:ab2} shows the effect of MLFM.
The results in the 1-3 rows indicate that when we use the concatenation as the fusion method, the performance improvement is marginal.
The main reason is that the insufficient fusion may introduce additional noisy information to the features, thereby limiting the performance improvement.
However, when we employ MLFM, there is a significant improvement compared with the absence of a fusion strategy.
As shown in Fig. \ref{fig:ab2}, the simple concatenation fusion strategy may confuse the model, leading to incorrect predictions.
In contrast, with our proposed MLFM, the model can better recognize the shape and contour of the entire objects.
This clearly demonstrates that our proposed MLFM ensures the full utilization of multi-level information from the SAM's encoder.

\noindent\textbf{Effectiveness of DEM.}
The 4-5 rows of Tab. \ref{tab:ab2} show the quantitative results with the DEM and its MEEM component.
As can be observed, the use of DEM without MEEM leads to a slight performance improvement.
However, the model still lacks of sufficient detail information.
Furthermore, when incorporating MEEM, the model results in a significant performance improvement.
These results demonstrate that the use of MEEM enables the model to capture more detail information.
Fig. \ref{fig:ab2} confirms that with DEM containing MEEM, the model obtains more fine-grained details and edges, achieving better SOD predictions.
\begin{table}[t]
\renewcommand{\arraystretch}{1.2}
\setlength{\tabcolsep}{1pt}
\centering
\caption{Ablation studies of our proposed modules. MLFB* indicates using $\mathrm{\textbf{F}}^{c}$ as the output of MLFM. DEM* indicates deleting MEEM. The best scores are marked in bold.}
\label{tab:ab2}
\resizebox{0.40\textwidth}{!}{
\begin{tabular}{c|ccc|ccc}
     \hline
     \multirow{2}{*}{\textbf{Method}}
     & \multicolumn{3}{c|}{\textbf{DUTS-TE}} & \multicolumn{3}{c}{\textbf{DUT-OMRON}}  \\
     & $MAE$ & $F^{max}_\beta$ & $S_m$ & $MAE$ & $F^{max}_\beta$ & $S_m$ \\
     \hline
     (a) Full fine-tuning & 0.030 & 0.921 & 0.909 & 0.044 & 0.865 & 0.869 \\

     (b) SAM+LMSA & 0.027 & 0.927 & 0.917 & 0.043 & 0.872 & 0.874  \\

     (c) SAM+LMSA+MLFM* & 0.027 & 0.928 & 0.918 & 0.042 & 0.871 & 0.873 \\

     (d) SAM+LMSA+MLFM &  0.025 & 0.931 & 0.920 & 0.041 & 0.876 & 0.876 \\

     (e) SAM+LMSA+MLFM+DEM* &  0.025 & 0.932 &\textbf{0.921} & 0.041 & 0.878 & 0.877 \\
     \hline
     (f) SAM+LMSA+MLFM+DEM &  \textbf{0.024} & \textbf{0.937} & 0.920 & \textbf{0.039} & \textbf{0.887} & \textbf{0.878} \\
     \hline
\end{tabular}
}
\end{table}
\section{Conclusion}
In this paper, we propose a novel feature learning framework named MDSAM for the SOD task.
The framework preserves the pre-trained weights of SAM, while incorporating multi-scale and fine-grained information.
Specifically, by introducing LMSA into SAM's encoder, we adapt SAM to SOD and enable the model to learn multi-scale information.
Furthermore, we propose the MLFM to fuse the output features at different layers of SAM's encoder effectively.
To enhance the SOD performance, we propose the DEM to address the issue of lacking fine-grained details in SAM.
Experimental results verify the effectiveness and strong generalization of our method.
\begin{acks}
This work was supported in part by the National Natural Science Foundation of China (No. 62101092) and the Fundamental Research Funds for the Central Universities (No. DUT23YG232).
\end{acks}

\clearpage
\bibliographystyle{ACM-Reference-Format}
\bibliography{sample-base}

\clearpage
\appendix
\begin{table*}[t]
\begin{center}
\caption{Performance comparison between our method and other SOD and COD methods on COD10K, CAMO and NC4K datasets. The best, second and third results are highlighted in {\color{red}{red}}, {\color{green}{green}}, and {\color{blue}{blue}}, respectively. \textbf{MDSAM} is with a $512 \times 512$ input. \textbf{MDSAM*} is with a $384 \times 384$ input.}
\label{tab:gen}
\resizebox{0.96\textwidth}{!}{
    \begin{tabular}{r|c|cccc|cccc|cccc}
      \hline
      \multirow{2}{*}{\textbf{Year}}
      & \multicolumn{1}{c|}{\multirow{2}{*}{\textbf{Method}}} & \multicolumn{4}{c|}{\textbf{COD10K}} & \multicolumn{4}{c|}{\textbf{NC4K}} & \multicolumn{4}{c}{\textbf{CAMO}}  \\

      && $MAE$ & $F^{m}_\beta$ & $S_m$ & $E_m$ &
      $MAE$ & $F^{m}_\beta$ & $S_m$ & $E_m$ &
      $MAE$ & $F^{m}_\beta$ & $S_m$ & $E_m$  \\
      \hline

      \multicolumn{14}{c}{\textbf{Salient Object Detection}} \\
      \hline

      2020 & F3Net~\cite{wei2020f3net}
      & 0.051 & 0.593 & 0.739 & 0.795
      & 0.070 & 0.689 & 0.767 & 0.793
      & 0.109 & 0.616 & 0.711 & 0.741\\

      2020 & MINet-R~\cite{pang2020multi}
      & 0.042 & 0.657 & 0.770 & 0.859
      & 0.056 & 0.764 & 0.812 & 0.887
      & 0.090 & 0.691 & 0.748 & 0.838\\

      2021 & VST~\cite{liu2021visual}
      & 0.042 & 0.653 & 0.781 & 0.837
      & 0.050 & 0.771 & 0.831 & 0.877
      & 0.076 & 0.738 & 0.787 & 0.838\\

      \hline
      \multicolumn{14}{c}{\textbf{Camouflaged Object Detection}} \\
      \hline

      2021 & MGL-R~\cite{zhai2021mutual}
      & 0.035 & 0.711 & 0.814 & 0.852
      & 0.052 & 0.782 & 0.833 & 0.867
      & 0.088 & 0.726 & 0.775 & 0.812\\

      2021 & C2FNet~\cite{sun2021context}
      & 0.036 & 0.723 & 0.813 & 0.890
      & 0.049 & 0.795 & 0.838 & 0.897
      & 0.080 & 0.762 & 0.796 & 0.854\\

      2021 & SINet-v2~\cite{fan2021concealed}
      & 0.037 & 0.718 & 0.815 & 0.887
      & 0.048 & 0.805 & 0.847 & 0.903
      & 0.070 & 0.782 & 0.820 & 0.882\\

      2022 & BSA-Net~\cite{zhu2022can}
      & 0.034 & 0.738 & 0.818 & 0.891
      & 0.048 & 0.808 & 0.841 & 0.897
      & 0.079 & 0.763 & 0.794 & 0.851\\

      2022 & BGNet~\cite{sun2022boundary}
      & 0.033 & 0.753 & 0.831 & \color{blue}0.901
      & 0.044 & 0.820 & 0.851 & 0.907
      & 0.073 & 0.789 & 0.812 & 0.870\\

      2022 & ZoomNet~\cite{pang2022zoom}
      & 0.029 & 0.766 & 0.838 & 0.888
      & 0.043 & 0.818 & 0.853 & 0.896
      & 0.066 & 0.794 & 0.820 & 0.878 \\

      2023 & FEDER~\cite{he2023camouflaged}
      & 0.031 & 0.751 & 0.822 & 0.900
      & 0.044 & 0.824 & 0.847 & 0.907
      & 0.071 & 0.781 & 0.802 & 0.867 \\

      2023 & FSPNet~\cite{huang2023feature}
      & \color{green}0.026 & \color{blue}0.769 & \color{green}0.851 & 0.895
      & \color{red}0.035 & \color{green}0.843 & \color{red}0.879 & \color{green}0.915
      & \color{red}0.050 & \color{green}0.830 & \color{red}0.856 & \color{green}0.899\\

      \hline
      2024 & \textbf{MDSAM*}
      & \color{blue}0.028 & \color{green}0.778 & \color{blue}0.839 & \color{green}0.905
      & \color{blue}0.040 & \color{blue}0.837 & \color{blue}0.864 & \color{blue}0.910
      & \color{blue}0.056 & \color{blue}0.822 & \color{blue}0.841 & \color{blue}0.888\\

      2024 & \textbf{MDSAM}
      & \color{red}0.025 & \color{red}0.803 & \color{red}0.862 & \color{red}0.921
      & \color{green}0.037 & \color{red}0.850 & \color{green}0.875 & \color{red}0.921
      & \color{green}0.053 & \color{red}0.834 & \color{green}0.852 & \color{red}0.903\\

      \hline
    \end{tabular}
}
\end{center}
\end{table*}
\begin{figure*}[t]
\centering
\includegraphics[scale=0.68]{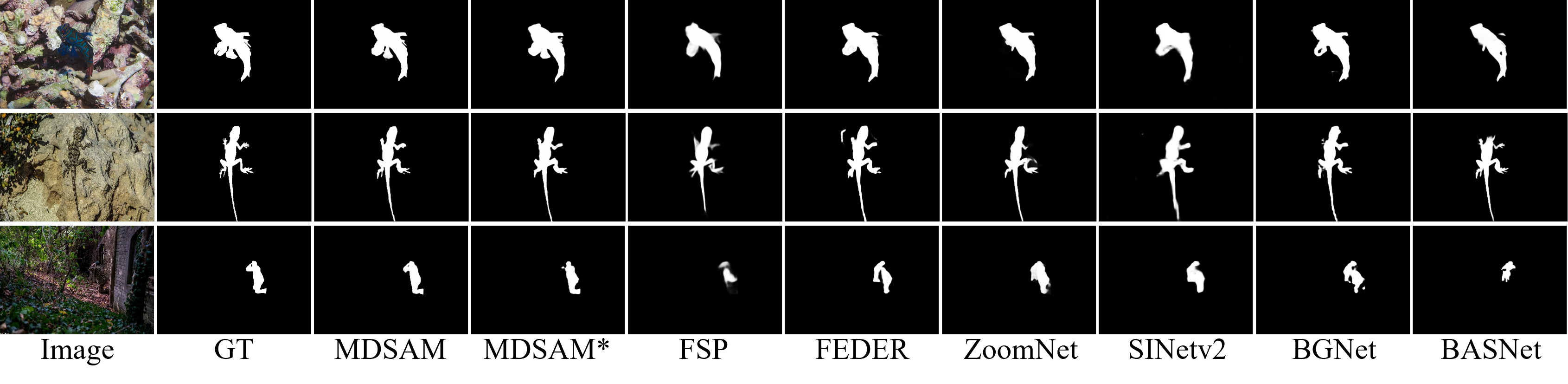}
\vspace{-4mm}
\caption{Visual comparison of our method and 6 other COD methods. \textbf{MDSAM} is with a $512 \times 512$ input resolution. \textbf{MDSAM*} is with a $384 \times 384$ input resolution.}
\label{fig:gen}
\end{figure*}
\section{Appendix}
\subsection{Model Generalization}
The original SAM exhibits strong generalization capabilities.
Our MDSAM not only introduces multi-scale and fine-grained details to SAM, but also retains the generalization capability.
To verify this, we conduct experiments for camouflaged object detection and polyp segmentation, which have very different domain characteristics.

\textbf{Camouflaged Object Detection (COD)}: Unlike SOD, which aims to find salient objects, COD focuses on detecting camouflaged objects in complex scenes.
Here, we adopt three COD datasets for model training and testing.
COD10K~\cite{fan2021concealed} contains 5,066 camouflaged, 1,934 non-camouflaged, 3000 background images.
CAMO~\cite{le2019anabranch} contains 1,250 camouflaged and 1,250 non-camouflaged images.
NC4K~\cite{lv2021simultaneously} contains 4,121 camouflaged images.
We employ a training strategy similar to~\cite{huang2023feature}.
We train our MDSAM with all images containing camouflaged objects in the COD10K training dataset and CAMO training datasets.
Then, we test MDSAM on all the test datasets.
Our MDSAM is compared with three SOD models (i.e., F3Net~\cite{wei2020f3net}, MINet~\cite{pang2020multi} and VST~\cite{liu2021visual}) and eight COD models (i.e., MGL-R~\cite{zhai2021mutual} C2FNet~\cite{sun2021context}, SINetv2~\cite{fan2021concealed}, BSA-Net~\cite{zhu2022can}, BGNet~\cite{sun2022boundary}, ZoomNet~\cite{pang2022zoom}, FEDER~\cite{he2023camouflaged} and FSPNet~\cite{huang2023feature}).
We adopt four metrics for evaluation.
Unlike SOD, we replaced the max F-measure ($F^{max}_\beta$) with the mean F-measure ($F^{m}_\beta$).
As shown in Tab. \ref{tab:gen}, our MDSAM achieves comparable performances on the COD task.
In qualitative evaluation, as shown in Fig. \ref{fig:gen}, our MDSAM demonstrates more precise localization and fine-grained details.

\textbf{Polyp Segmentation}: We further conduct experiments on polyp segmentation, which is a typical medical image segmentation task.
We employ the same training strategy as Poly-PVT~\cite{dong2021polyp}, and test our method on Kvasir \cite{jha2020kvasir}, CVC-300 \cite{bernal2012towards}.
As shown in the Tab. \ref{tab:ps}, our method achieves superior performance on polyp segmentation.
Furthermore, compared with Adapter-enhanced SAM, our method is much better on both COD and polyp segmentation.

From the above experiments, one can see that our MDSAM not only performs excellently on SOD tasks but also exhibits outstanding performance on COD and Polyp Segmentation.
They clearly demonstrate the excellent generalization capabilities of our model.
\begin{table}[t]
\begin{center}
\caption{Experiments on COD and polyp segmentation.}
\label{tab:ps}
\resizebox{0.48\textwidth}{!}{
    \begin{tabular}{c|cccc|cccc}
      \hline
      \multirow{3}{*}{\textbf{Method}}
      & \multicolumn{4}{c|}{\textbf{COD}} & \multicolumn{4}{c}{\textbf{Polyp Segmentation}} \\
      & \multicolumn{2}{c}{\textbf{COD10k}} & \multicolumn{2}{c|}{\textbf{NC4K}}
      & \multicolumn{2}{c}{\textbf{Kvasir}} & \multicolumn{2}{c}{\textbf{CVC-300}}\\
      & $MAE$ & $E_m$ & $MAE$ & $E_m$ & $MAE$ & $E_m$ & $MAE$ & $E_m$ \\
      \hline
      Polyp-PVT\cite{dong2021polyp} & - & - & - & - & 0.023 & 0.956 & 0.011 & 0.961\\
      SAM (Adapter) & 0.028 & 0.913 & 0.040  & 0.912 & 0.025 & 0.954 & 0.010 & 0.962\\
      \hline
      MDSAM  & 0.025 & 0.921 & 0.037 & 0.921 & 0.021 & 0.961 & 0.008 & 0.964\\
      \hline
    \end{tabular}
}
\end{center}
\end{table}
\subsection{Zero-shot Analysis}
SAM has a strong segmentation ability when prompts are provided. 
We extract bounding boxes from each connected component in the ground truth and used it as box prompts for SAM. 
We also conduct experiments with SAM-HQ~\cite{ke2024segment}, which enables SAM to segment high-quality results. 
As shown in the Tab. \ref{tab:zero}, when accurate prompts are given, SAM performs significantly well on SOD. 
However, SOD requires corresponding semantic information. 
SAM performs poorly when no prompts are given. 
Our MDSAM not only improves upon SAM's segmentation but also learns the semantic information of SOD.
\begin{table}[t]
\begin{center}
\caption{Zero-shot performance comparisons on SOD.}
\label{tab:zero}
\resizebox{0.48\textwidth}{!}{
    \begin{tabular}{c|cc|cc|cc}
      \hline
      \multirow{2}{*}{\textbf{Method}}
      & \multicolumn{2}{c|}{\textbf{DUTS-TE}} & \multicolumn{2}{c|}{\textbf{DUT-OMRON}}
      & \multicolumn{2}{c}{\textbf{HKU-IS}} \\
      & $MAE$ & $F^{max}_\beta$
      & $MAE$ & $F^{max}_\beta$
      & $MAE$ & $F^{max}_\beta$\\
      \hline
      SAM (Zero-shot) & 0.162 & 0.179 & 0.163 & 0.182 & 0.197 & 0.232 \\
      SAM-HQ (Zero-shot) & 0.148 & 0.200 & 0.147 & 0.191 & 0.190 & 0.243\\
      SAM (Box) & 0.015 & 0.949 & 0.014 & 0.952 & 0.014 & 0.968\\
      SAM-HQ (Box) & 0.015 & 0.949 & 0.014 & 0.951 & 0.015 & 0.967 \\
      \hline
      MDSAM  & 0.024 & 0.937 & 0.039 & 0.887 & 0.019 & 0.963\\
      \hline
    \end{tabular}
}
\end{center}
\end{table}
\subsection{More Ablation Studies}
In our proposed LMSA, we use Average Pooling (AP) to obtain multi-scale features. 
In addition, we introduce local-detail information to address the shortcomings of SAM. 
In this section, we conduct experiments on varied scales of LMSA and whether to introduce local information in MDSAM.
The experiments are conducted under the fully designed MDSAM and the input resolution of the model is set to $512 \times 512$.

\begin{figure}[t]
\centering
\includegraphics[scale=0.3]{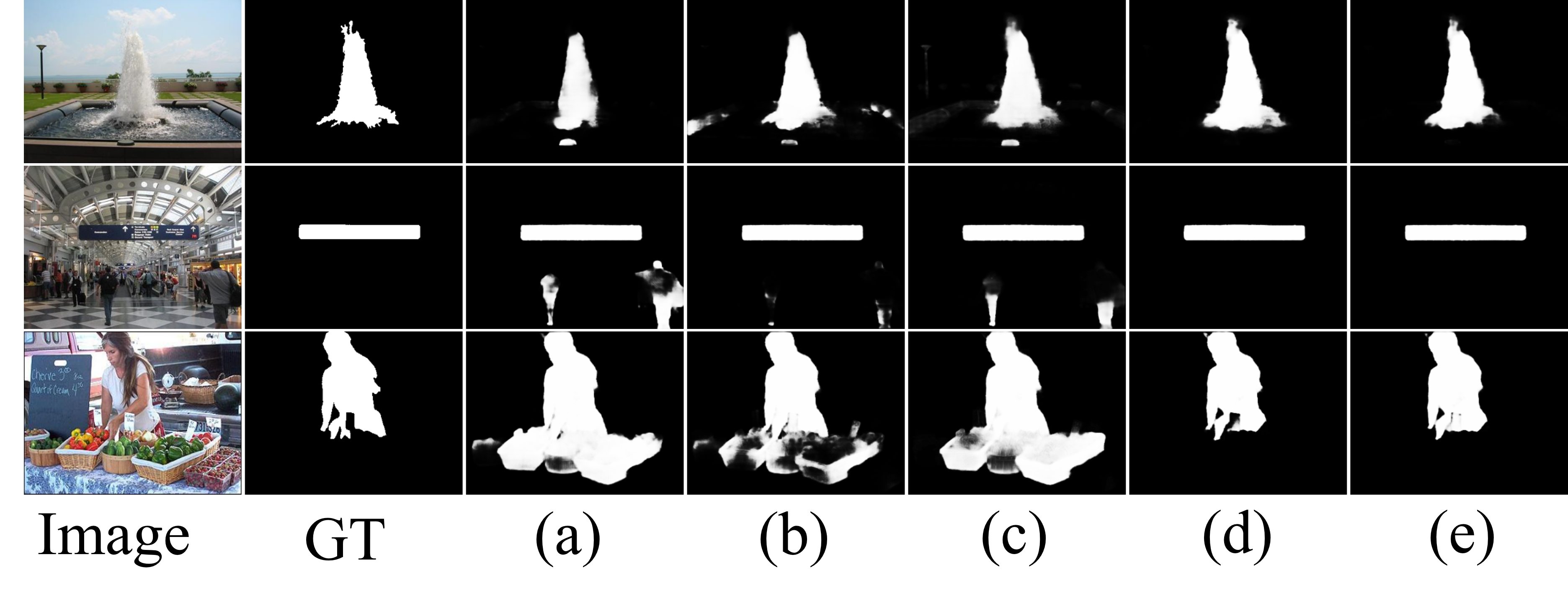}
\caption{Visual comparison of different scales and local information.}
\label{fig:mab}
\end{figure}

\textbf{Effects of Scales.} 
Tab.~\ref{tab:msb} shows the effect of using different pooling scales. 
As indicated by methods (a) and (b), as well as (d) and (e), the change in scales has a very small impact on the results as long as it is multi-scale.
However, as shown in methods (c) and (e), multi-scale methods usually perform better than single-scale methods. 
Fig.~\ref{fig:mab} demonstrates that, compared with the single-scale setting (c), the multi-scale settings (d) and (e) can more accurately detect salient objects in complex scenes. 
Therefore, the scale setting of LMSA prefers to maintain multi-scale.

\textbf{Effectiveness of Local Information.} 
To further explore the effect of local information, we conduct more experiments presented in Tab.~\ref{tab:msb}.
It can be observed that with the introduction of local information, the model's performance has significantly improved.
Additionally, when compared with method (b) and method (e), one can see that our MDSAM exhibits better performances with the same scales under the presence of local information.
Fig.~\ref{fig:mab} illustrates that local information enables the model to extract more precise features, resulting in better saliency maps.
\begin{table}[t]
\renewcommand{\arraystretch}{1.2}
\setlength{\tabcolsep}{1pt}
\centering
\caption{Results of different scales and local information.}
\label{tab:msb}
\resizebox{0.45\textwidth}{!}{
\begin{tabular}{c|c|c|ccc|ccc}
     \hline
     \multirow{2}{*}{\textbf{Method}} & \multirow{2}{*}{\textbf{Scale}} &\multirow{2}{*}{\textbf{Local}}
     & \multicolumn{3}{c|}{\textbf{DUTS-TE}} & \multicolumn{3}{c}{\textbf{DUT-OMRON}}  \\
     &&& $MAE$ & $F^{max}_\beta$ & $S_m$ & $MAE$ & $F^{max}_\beta$ & $S_m$ \\
     \hline
     (a) &1,2,3,6 & $\times$ & 0.026 & 0.927 & 0.910 & 0.044 & 0.865 & 0.869 \\

     (b) & 3,6,9,12 & $\times$& 0.026 & 0.928 & 0.912 & 0.042 & 0.872 & 0.873  \\

     (c) & 9,9,9,9 &$\checkmark$& 0.025 & 0.930 & 0.916 & 0.041 & 0.875 & 0.873 \\

     (d) &3,5,7,9 &$\checkmark$&  \textbf{0.024} & 0.936& \textbf{0.921} & \textbf{0.039} & 0.882 & \textbf{0.881} \\
     \hline
     (e) \textbf{MDSAM} &3,6,9,12&$\checkmark$&  \textbf{0.024} & \textbf{0.937} & 0.920 & \textbf{0.039} & \textbf{0.887} & 0.878 \\
     \hline
\end{tabular}
}
\end{table}
\subsection{Attribute-based Analysis}
\begin{figure*}[t!p]
\centering
\includegraphics[scale=0.65]{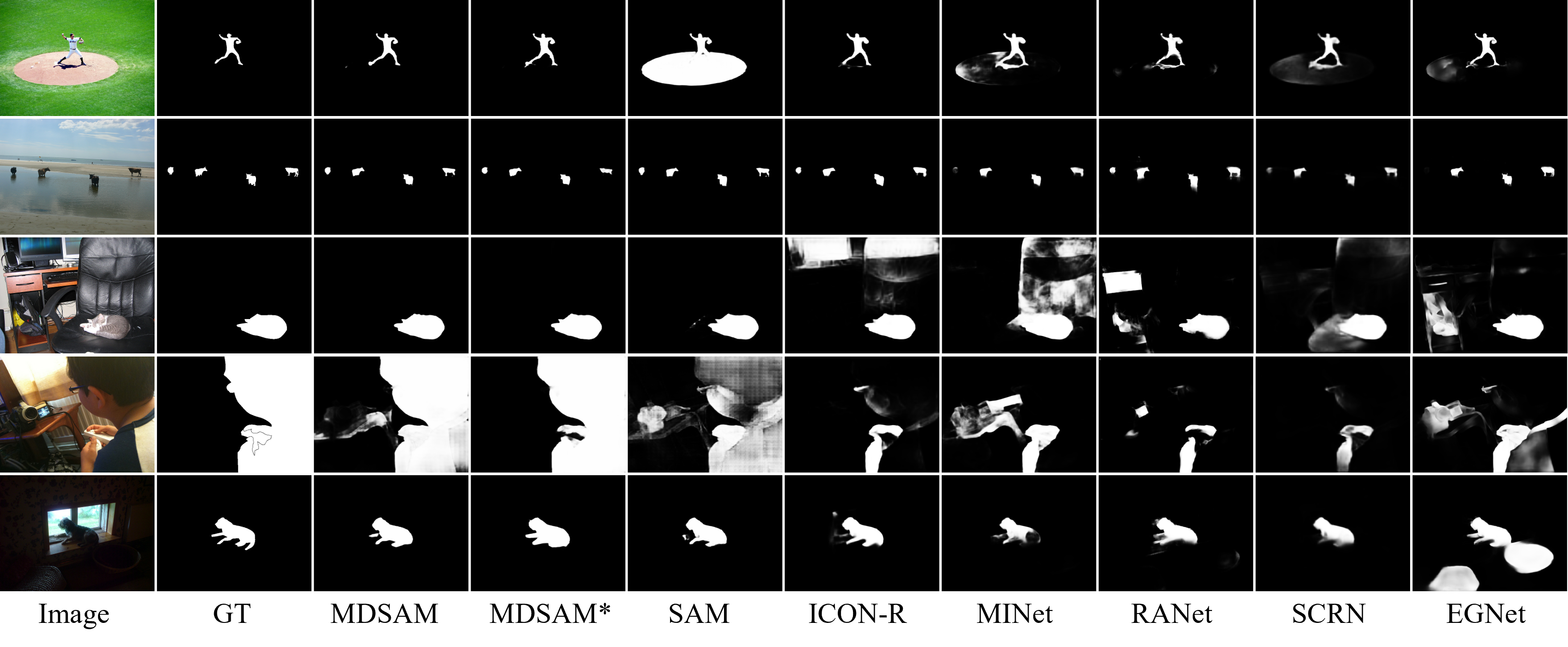}
\caption{Visual comparison of saliency maps with our model and 6 representative methods. MDSAM is with a $512 \times 512$ input resolution. MDSAM* is with a $384 \times 384$ input resolution.}
\label{fig:soc1}
\end{figure*}

In this section, we provide an attribute-based analysis by evaluating our proposed method on the challenging SOC~\cite{fan2022salient} dataset. 
The SOC test dataset is divided into 9 major categories, which are Appearance Change (AC, 79 images), Big Object (BO, 24 images), Clutter (CL, 92 images), Heterogeneous Object (HO, 153 images), Motion Blur (MB, 32 images), Occlusion (OC, 157 images), Out-of-View (OV, 155 images), Shape Complexity (SC, 116 images), and Small Object (SO, 389 images). 
We compare our MDSAM with 17 methods, including Amulet~\cite{zhang2017amulet}, DSS~\cite{hou2017deeply}, NLDF~\cite{luo2017non}, SRM~\cite{wang2017stagewise}, BMPM~\cite{zhang2018bi}, C2SNet~\cite{li2018contour}, DGRL~\cite{wang2018detect}, RANet~\cite{chen2020reverse}, CPD~\cite{wu2019cascaded}, EGNet~\cite{zhao2019egnet}, PoolNet~\cite{liu2019simple}, SCRN~\cite{wu2019stacked}, BANet~\cite{su2019selectivity}, MINet~\cite{pang2020multi}, ICON-R~\cite{zhuge2022salient}, DC-Net-R~\cite{zhu2023dc}, and full fine-tuned SAM~\cite{kirillov2023segment}.
As shown in Tab.~\ref{tab:soc}, our MDSAM demonstrates superior performance in most scenarios at resolutions of $512 \times 512$ and $384 \times 384$. 
And they perform averagely only in the categories AC and BO, which have a small amount of data. 
Fig.~\ref{fig:soc1} shows some visualization results of the proposed MDSAM and six representative state-of-the-art methods.
This visualization demonstrates that current methods struggle to accurately localize both large and small objects, and the results lack fine-grained details.
Our MDSAM can accurately locate multi-scale objects, and both edges and details are highly precise.
\begin{table*}[t]
\begin{center}
\caption{Performance comparison between our method and other 17 methods on the SOC dataset in terms of $MAE$, $F^{\omega}_\beta$, $S_m$, and $E_m$. The best, second and third results are highlighted in {\color{red}{red}}, {\color{green}{green}}, and {\color{blue}{blue}}, respectively. \textbf{MDSAM} is with a $512 \times 512$ input. \textbf{MDSAM*} is with a $384 \times 384$ input.}
\label{tab:soc}
\resizebox{\textwidth}{!}{
    \begin{tabular}{c|c|ccccccccccccccccc|cc}
      \hline
      \multirow{2}{*}{\textbf{Attr}}
      & \multirow{2}{*}{\textbf{Metrics}}
      & Amulet & DSS & NLDF & SRM & BMPM & C2SNet & DGRL & RANet & CPD
      & EGNet & PoolNet & SCRN & BANet & MINet & ICON-R & DC-Net-R & SAM
      & \multirow{2}{*}{\textbf{MDSAM*}} & \multirow{2}{*}{\textbf{MDSAM}} \\

      &&~\cite{zhang2017amulet} &~\cite{hou2017deeply} &~\cite{luo2017non} &~\cite{wang2017stagewise} &~\cite{zhang2018bi}
      &~\cite{li2018contour} &~\cite{wang2018detect} &~\cite{chen2020reverse} &~\cite{wu2019cascaded} &~\cite{zhao2019egnet}
      &~\cite{liu2019simple} &~\cite{wu2019stacked} &~\cite{su2019selectivity} &~\cite{pang2020multi}  &~\cite{zhuge2022salient} &~\cite{zhu2023dc}
      &~\cite{kirillov2023segment}
      &&\\
      \hline

      \multirow{4}{*}{\textbf{AC}} & $MAE$
       & 0.120 & 0.113 & 0.119 & 0.096 & 0.098 & 0.109 & 0.081 & 0.132 & 0.089 & 0.085 & 0.093 & \color{blue}0.078 & 0.086 & 0.079 & \color{red}0.062 & \color{green}0.076 & \color{blue}0.078 & 0.079 & \color{blue}0.078
      \\
       & $F^{\omega}_\beta$
       & 0.620 & 0.629 & 0.620 & 0.690 & 0.680 & 0.647 & 0.718 & 0.603 & 0.721 & 0.731 & 0.713 & 0.723 & 0.739 & \color{red}0.930 & \color{green}0.784 & \color{blue}0.768 & 0.757 & 0.759 & 0.764
       \\
       & $S_m$
       & 0.752 & 0.753 & 0.737 & 0.791 & 0.780 & 0.755 & 0.791 & 0.709 & 0.799 & 0.806 & 0.795 & 0.809 & 0.806 & 0.802 & \color{red}0.835 & \color{green}0.824 & \color{blue}0.821 & 0.815 & 0.819
      \\
       & $E_m$
       & 0.790 & 0.787 & 0.793 & 0.824 & 0.815 & 0.806 & 0.853 & 0.765 & 0.852 & 0.854 & 0.846 & 0.848 & 0.858 & 0.843 & 0.891 & 0.867 & 0.860 & 0.863 & 0.868
      \\
      \hline

      \multirow{4}{*}{\textbf{BO}} & $MAE$
       & 0.334 & 0.343 & 0.341 & 0.294 & 0.292 & 0.257 & 0.207 & 0.440 & 0.236 & 0.358 & 0.339 & 0.217 & 0.261 & 0.175 & 0.200 & 0.278 & 0.232 & 0.264 & 0.231
      \\
       & $F^{\omega}_\beta$
       & 0.625 & 0.628 & 0.635 & 0.679 & 0.683 & 0.739 & \color{green}0.794 & 0.469 & 0.755 & 0.602 & 0.625 & 0.784 & 0.729 & \color{red}0.828 & \color{green}0.794 & 0.699 & 0.749 & 0.709 & 0.758
      \\
       & $S_m$
       & 0.589 & 0.577 & 0.583 & 0.628 & 0.619 & 0.667 & 0.696 & 0.437 & 0.679 & 0.546 & 0.578 & \color{blue}0.707 & 0.657 & \color{red}0.743 & \color{green}0.714 & 0.637 & 0.684 & 0.658 & 0.687
      \\
       & $E_m$
       & 0.566 & 0.554 & 0.556 & 0.630 & 0.635 & 0.674 & \color{blue}0.736 & 0.423 & 0.699 & 0.547 & 0.572 & 0.716 & 0.663 & \color{red}0.769 & \color{green}0.740 & 0.641 & 0.703 & 0.681 & 0.709
      \\
      \hline

      \multirow{4}{*}{\textbf{CL}} & $MAE$
       & 0.141 & 0.153 & 0.159 & 0.134 & 0.123 & 0.144 & 0.119 & 0.188 & 0.112 & 0.139 & 0.134 & 0.113 & 0.117 & 0.108 & 0.113 & 0.112 & \color{blue}0.098 & \color{green}0.092 & \color{red}0.090
      \\
       & $F^{\omega}_\beta$
       & 0.763 & 0.721 & 0.713 & 0.758 & 0.760 & 0.742 & 0.769 & 0.633 & \color{green}0.786 & 0.757 & 0.760 & \color{red}0.795 & \color{blue}0.784 & 0.719 & 0.736 & 0.746 & 0.772 & 0.772 & 0.777
      \\
       & $S_m$
       & 0.763 & 0.721 & 0.713 & 0.758 & 0.760 & 0.742 & 0.769 & 0.633 & 0.786 & 0.757 & 0.760 & 0.795 & 0.784 & 0.783 & 0.791 & 0.798 & \color{blue}0.818 & \color{green}0.819 & \color{red}0.823
      \\
       & $E_m$
       & 0.788 & 0.763 & 0.764 & 0.792 & 0.801 & 0.789 & 0.824 & 0.715 & 0.823 & 0.789 & 0.800 & 0.819 & 0.824 & 0.819 & 0.832 & 0.834 & \color{blue}0.847 & \color{green}0.851 & \color{red}0.855
      \\
      \hline

      \multirow{4}{*}{\textbf{HO}} & $MAE$
       & 0.119 & 0.124 & 0.126 & 0.115 & 0.116 & 0.123 & 0.104 & 0.143 & 0.098 & 0.106 & 0.100 & 0.096 & 0.094 & 0.089 & 0.091 & 0.092 & \color{blue}0.087 & \color{green}0.083 & \color{red}0.080
      \\
       & $F^{\omega}_\beta$
       & 0.688 & 0.660 & 0.661 & 0.696 & 0.684 & 0.668 & 0.722 & 0.626 & 0.736 & 0.720 & 0.739 & 0.743 & 0.753 & 0.759 & 0.767 & 0.761 & \color{green}0.793 & \color{blue}0.788 & \color{red}0.795
      \\
       & $S_m$
       & 0.790 & 0.767 & 0.755 & 0.794 & 0.781 & 0.768 & 0.791 & 0.713 & 0.807 & 0.802 & 0.815 & 0.823 & 0.819 & 0.821 & 0.823 & 0.818 & \color{green}0.837 & \color{blue}0.835 & \color{red}0.842
      \\
       & $E_m$
       & 0.809 & 0.796 & 0.798 & 0.819 & 0.813 & 0.805 & 0.833 & 0.777 & 0.838 & 0.829 & 0.845 & 0.842 & 0.850 & 0.858 & \color{blue}0.865 & 0.848 & 0.864 & \color{green}0.869 & \color{red}0.872
      \\
      \hline

      \multirow{4}{*}{\textbf{MB}} & $MAE$
       & 0.142 & 0.132 & 0.138 & 0.115 & 0.105 & 0.128 & 0.113 & 0.139 & 0.104 & 0.109 & 0.121 & 0.100 & 0.104 & 0.105 & 0.100 & 0.109 & \color{blue}0.091 & \color{green}0.074 & \color{red}0.065
      \\
       & $F^{\omega}_\beta$
       & 0.561 & 0.577 & 0.551 & 0.619 & 0.651 & 0.593 & 0.655 & 0.576 & 0.655 & 0.649 & 0.642 & 0.690 & 0.670 & 0.676 & 0.699 & 0.676 & \color{green}0.758 & \color{blue}0.754 & \color{red}0.779
      \\
       & $S_m$
       & 0.712 & 0.719 & 0.685 & 0.742 & 0.762 & 0.719 & 0.744 & 0.696 & 0.753 & 0.762 & 0.751 & 0.792 & 0.764 & 0.761 & 0.774 & 0.757 & \color{blue}0.810 & \color{green}0.820 & \color{red}0.834
      \\
       & $E_m$
       & 0.762 & 0.760 & 0.755 & 0.780 & 0.799 & 0.784 & 0.808 & 0.718 & 0.818 & 0.798 & 0.800 & 0.800 & 0.808 & 0.821 & 0.828 & 0.787 & \color{blue}0.845 & \color{green}0.859 & \color{red}0.870
      \\
      \hline

      \multirow{4}{*}{\textbf{OC}} & $MAE$
       & 0.143 & 0.144 & 0.149 & 0.129 & 0.119 & 0.130 & 0.116 & 0.169 & 0.106 & 0.121 & 0.118 & 0.111 & 0.112 & 0.102 & 0.106 & 0.102 & \color{green}0.089 & \color{red}0.085 & \color{green}0.089
      \\
       & $F^{\omega}_\beta$
       & 0.607 & 0.595 & 0.593 & 0.630 & 0.644 & 0.622 & 0.659 & 0.527 & 0.679 & 0.658 & 0.659 & 0.673 & 0.677 & 0.686 & 0.683 & 0.708 & \color{green}0.732 & \color{blue}0.728 & \color{red}0.733
      \\
       & $S_m$
       & 0.735 & 0.719 & 0.709 & 0.749 & 0.752 & 0.738 & 0.747 & 0.641 & 0.773 & 0.754 & 0.756 & 0.775 & 0.766 & 0.771 & 0.771 & 0.787 & \color{red}0.808 & \color{blue}0.798 & \color{green}0.802
      \\
       & $E_m$
       & 0.762 & 0.760 & 0.755 & 0.780 & 0.799 & 0.784 & 0.808 & 0.718 & 0.818 & 0.798 & 0.800 & 0.800 & 0.808 & 0.821 & 0.817 & 0.824 & \color{blue}0.843 & \color{red}0.847 & \color{green}0.845
      \\
      \hline

      \multirow{4}{*}{\textbf{OV}} & $MAE$
       & 0.173 & 0.180 & 0.184 & 0.150 & 10.360 & 0.159 & 0.125 & 0.217 & 0.125 & 0.146 & 0.148 & 0.126 & 0.119 & 0.117 & 0.120 & 0.126 & \color{blue}0.110 & \color{green}0.100 & \color{red}0.097
      \\
       & $F^{\omega}_\beta$
       & 0.637 & 0.622 & 0.616 & 0.682 & 0.701 & 0.671 & 0.733 & 0.529 & 0.724 & 0.707 & 0.697 & 0.723 & 0.751 & 0.738 & 0.749 & 0.738 & \color{blue}0.769 & \color{green}0.780 & \color{red}0.790
      \\
       & $S_m$
       & 0.721 & 0.700 & 0.688 & 0.745 & 0.751 & 0.728 & 0.762 & 0.611 & 0.765 & 0.752 & 0.747 & 0.774 & 0.779 & 0.775 & 0.779 & 0.771 & \color{blue}0.796 & \color{green}0.804 & \color{red}0.811
      \\
       & $E_m$
       & 0.750 & 0.737 & 0.736 & 0.778 & 0.806 & 0.789 & 0.828 & 0.664 & 0.809 & 0.802 & 0.795 & 0.807 & 0.835 & 0.822 & 0.834 & 0.814 & \color{blue}0.842 & \color{green}0.857 & \color{red}0.860
      \\
      \hline

      \multirow{4}{*}{\textbf{SC}} & $MAE$
       & 0.098 & 0.098 & 0.101 & 0.090 & 0.081 & 0.100 & 0.087 & 0.110 & 0.076 & 0.083 & 0.075 & 0.078 & 0.078 & 0.077 & 0.080 & 0.072 & \color{blue}0.068 & \color{red}0.062 & \color{green}0.065
      \\
       & $F^{\omega}_\beta$
       & 0.608 & 0.599 & 0.593 & 0.638 & 0.677 & 0.611 & 0.669 & 0.594 & 0.701 & 0.678 & 0.695 & 0.691 & 0.705 & 0.711 & 0.714 & \color{blue}0.749 & \color{green}0.757 & 0.746 & \color{red}0.772
      \\
       & $S_m$
       & 0.768 & 0.761 & 0.746 & 0.783 & 0.799 & 0.756 & 0.772 & 0.724 & 0.807 & 0.793 & 0.807 & 0.809 & 0.807 & 0.808 & 0.808 & \color{blue}0.826 & \color{green}0.831 & 0.825 & \color{red}0.839
      \\
       & $E_m$
       & 0.793 & 0.798 & 0.787 & 0.812 & 0.840 & 0.805 & 0.837 & 0.791 & 0.848 & 0.843 & 0.856 & 0.843 & 0.850 & 0.859 & 0.871 & 0.869 & 0.866 & 0.882 & 0.882
      \\
      \hline

      \multirow{4}{*}{\textbf{SO}} & $MAE$
       & 0.119 & 0.109 & 0.115 & 0.099 & 0.096 & 0.116 & 0.092 & 0.113 & 0.084 & 0.098 & 0.087 & 0.082 & 0.091 & 0.820 & \color{blue}0.079 & 0.084 & \color{blue}0.079 & \color{green}0.072 & \color{red}0.070
      \\
       & $F^{\omega}_\beta$
       & 0.523 & 0.524 & 0.526 & 0.561 & 0.567 & 0.531 & 0.602 & 0.518 & 0.613 & 0.594 & 0.626 & 0.614 & 0.617 & 0.624 & 0.660 & 0.656 & \color{green}0.685 & \color{blue}0.676 & \color{red}0.694
      \\
       & $S_m$
       & 0.718 & 0.713 & 0.703 & 0.737 & 0.732 & 0.707 & 0.736 & 0.682 & 0.756 & 0.749 & 0.768 & 0.767 & 0.755 & 0.759 & 0.778 & 0.774 & \color{green}0.791 & \color{blue}0.786 & \color{red}0.798
      \\
       & $E_m$
       & 0.744 & 0.755 & 0.747 & 0.769 & 0.779 & 0.751 & 0.802 & 0.758 & 0.806 & 0.784 & 0.814 & 0.796 & 0.801 & 0.806 & \color{blue}0.835 & 0.813 & 0.834 & \color{green}0.844 & \color{red}0.847
      \\
      \hline
    \end{tabular}
}
\end{center}
\end{table*}
\subsection{More Comparison Results}
\begin{figure*}[t!p]
\centering
\includegraphics[scale=1]{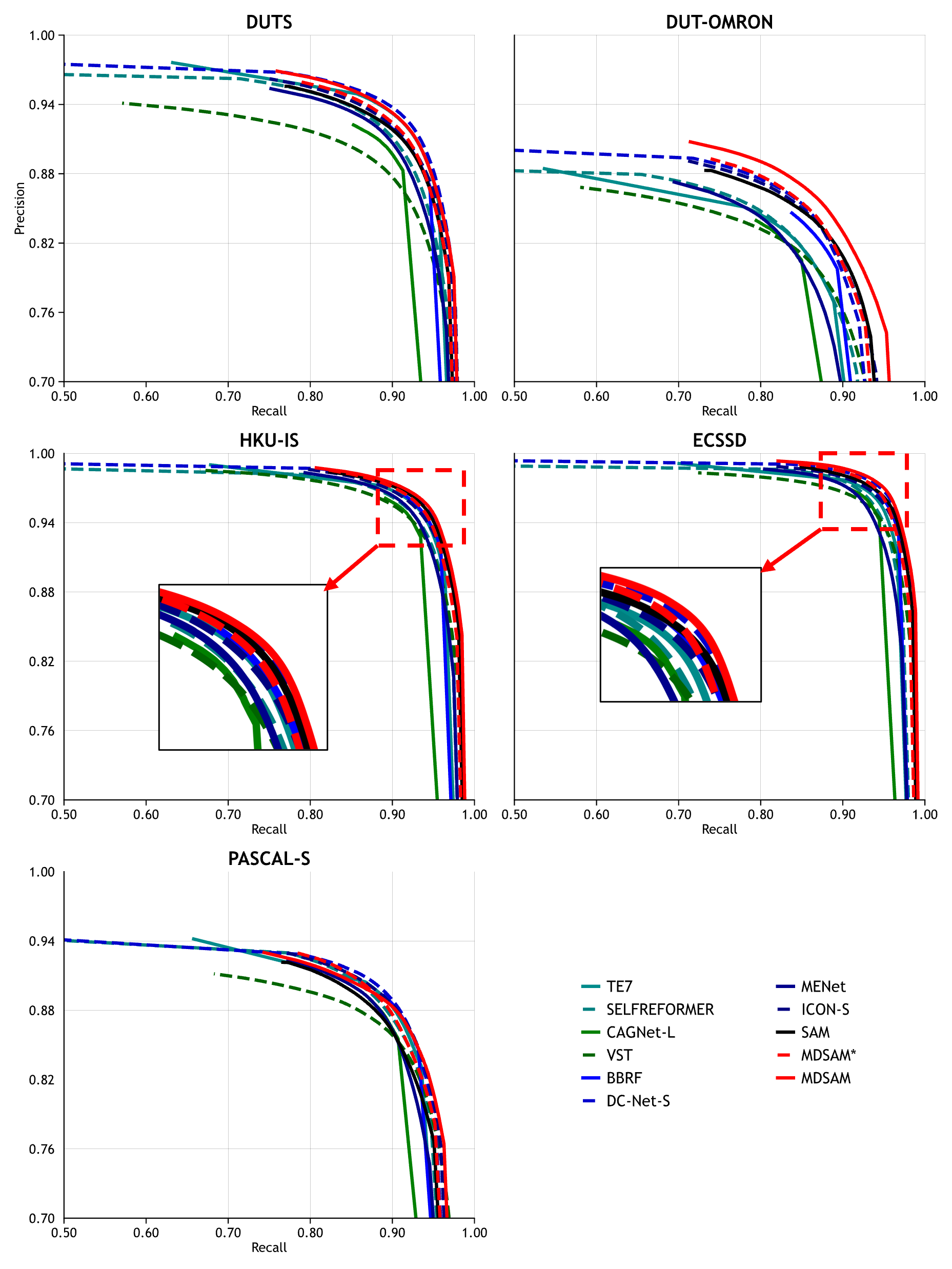}
\caption{Precision-Recall curves comparison on five SOD datasets.}
\label{fig:sod_pr}
\end{figure*}

\begin{figure*}[t!p]
\centering
\includegraphics[scale=1]{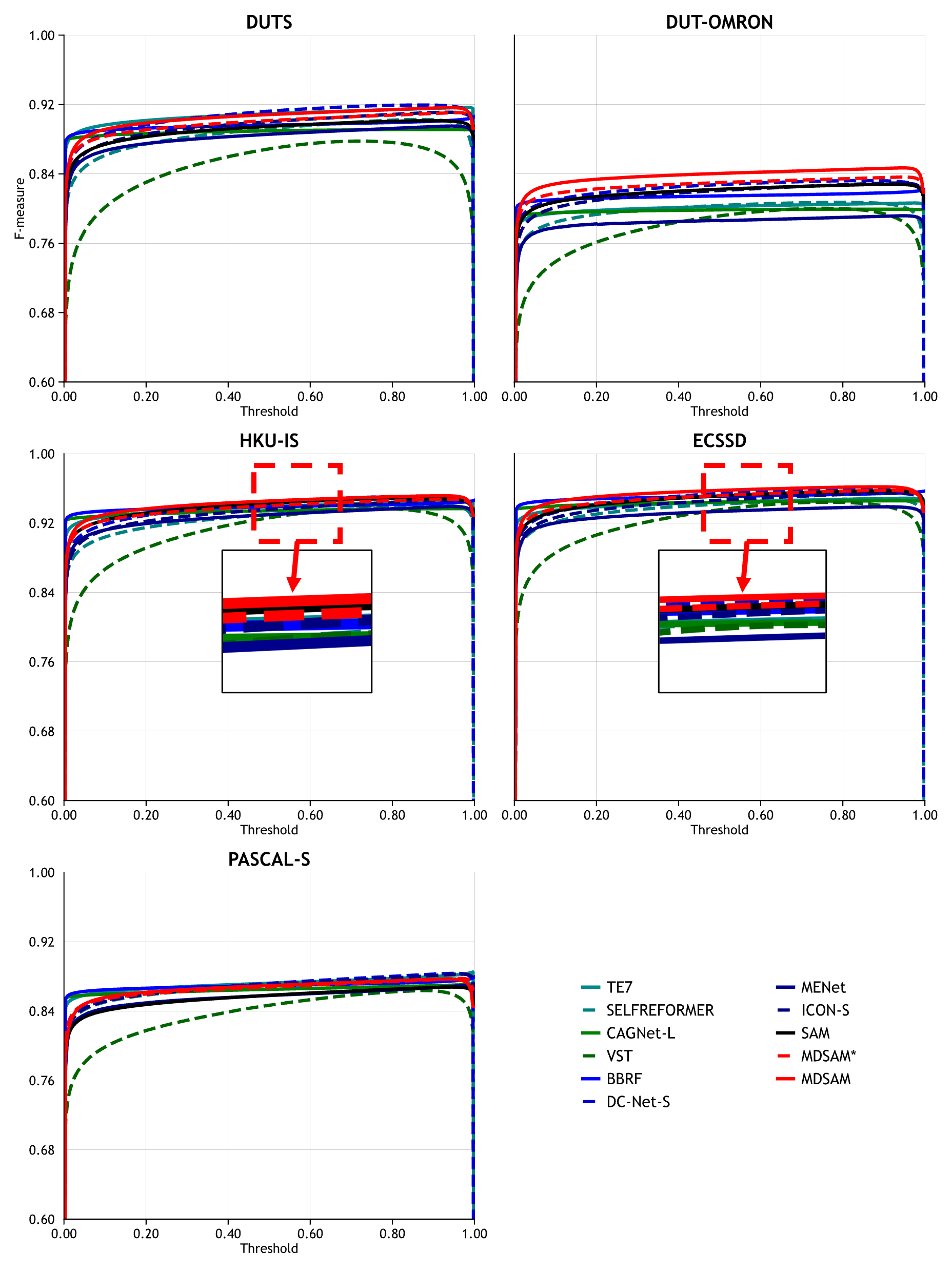}
\caption{F-measure curves comparison on five SOD datasets.}
\label{fig:sod_fm}
\end{figure*}

\begin{figure*}[t!p]
\centering
\includegraphics[scale=1]{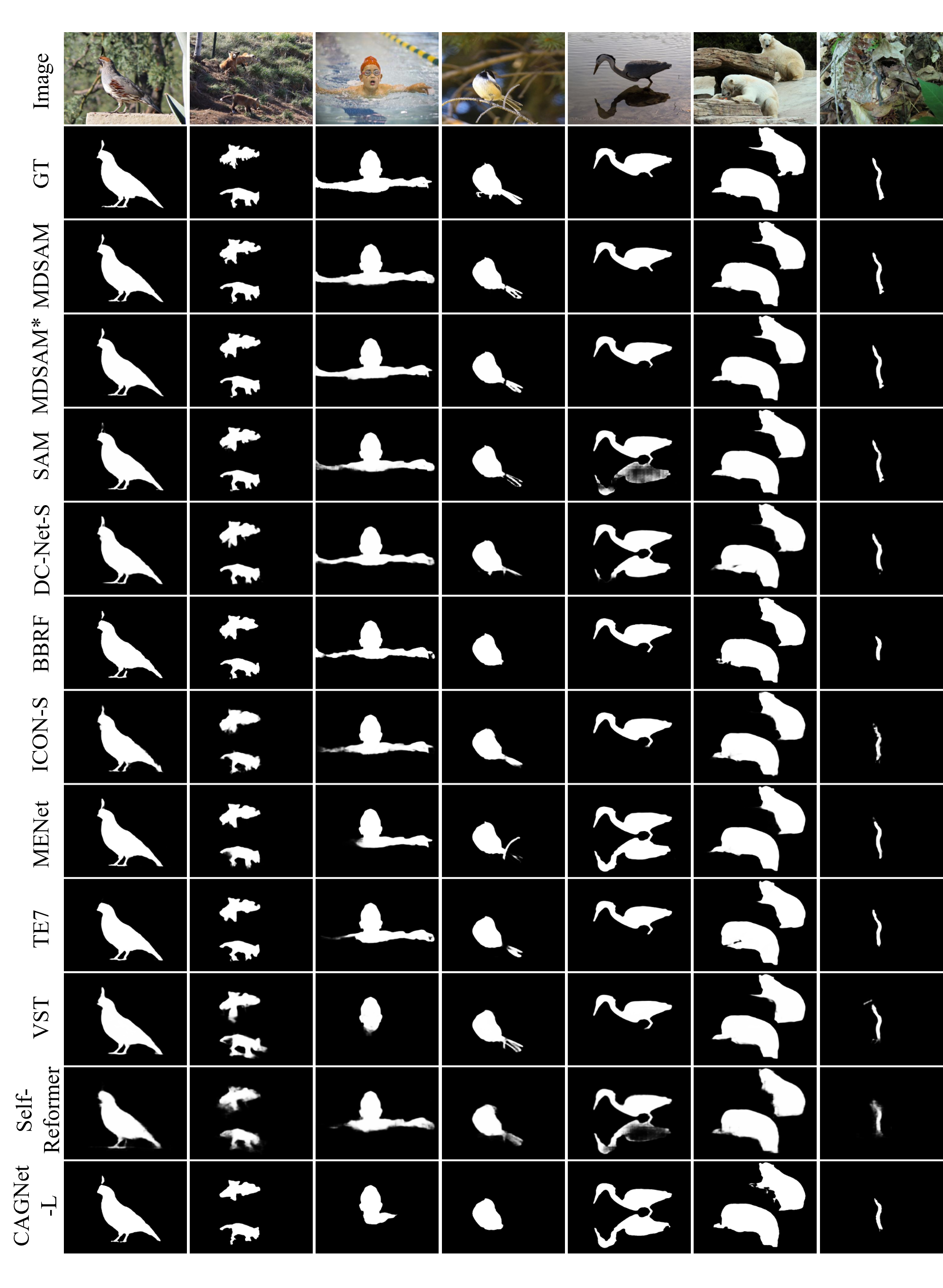}
\caption{Visual comparison of saliency maps with our model and 9 representative methods on five SOD datasets.}
\label{fig:sp1}
\end{figure*}

\begin{figure*}[t!p]
\centering
\includegraphics[scale=1]{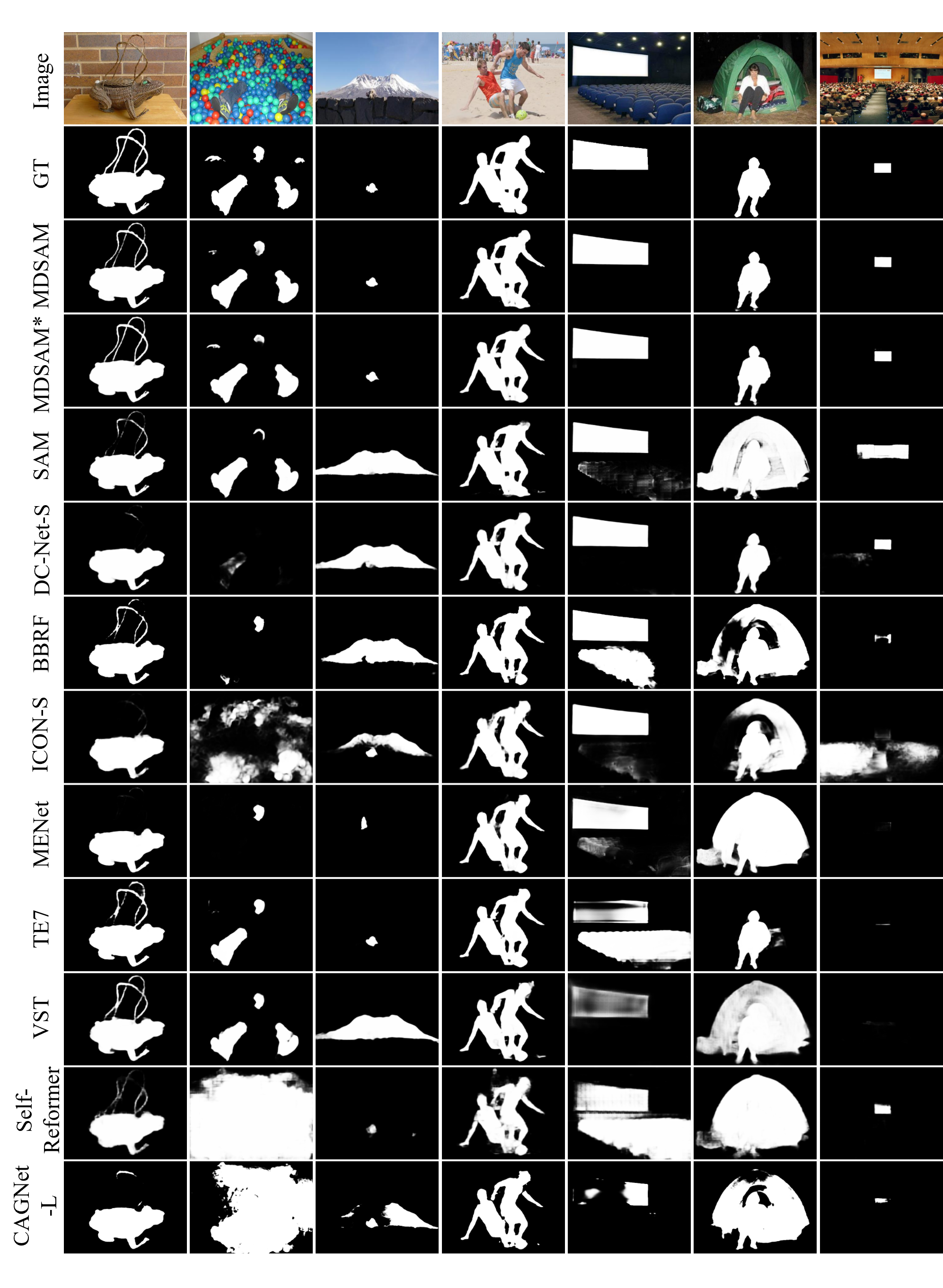}
\caption{Visual comparison of saliency maps with our model and 9 representative methods on five SOD datasets.}
\label{fig:sp2}
\end{figure*}

\begin{figure*}[t!p]
\centering
\includegraphics[scale=1]{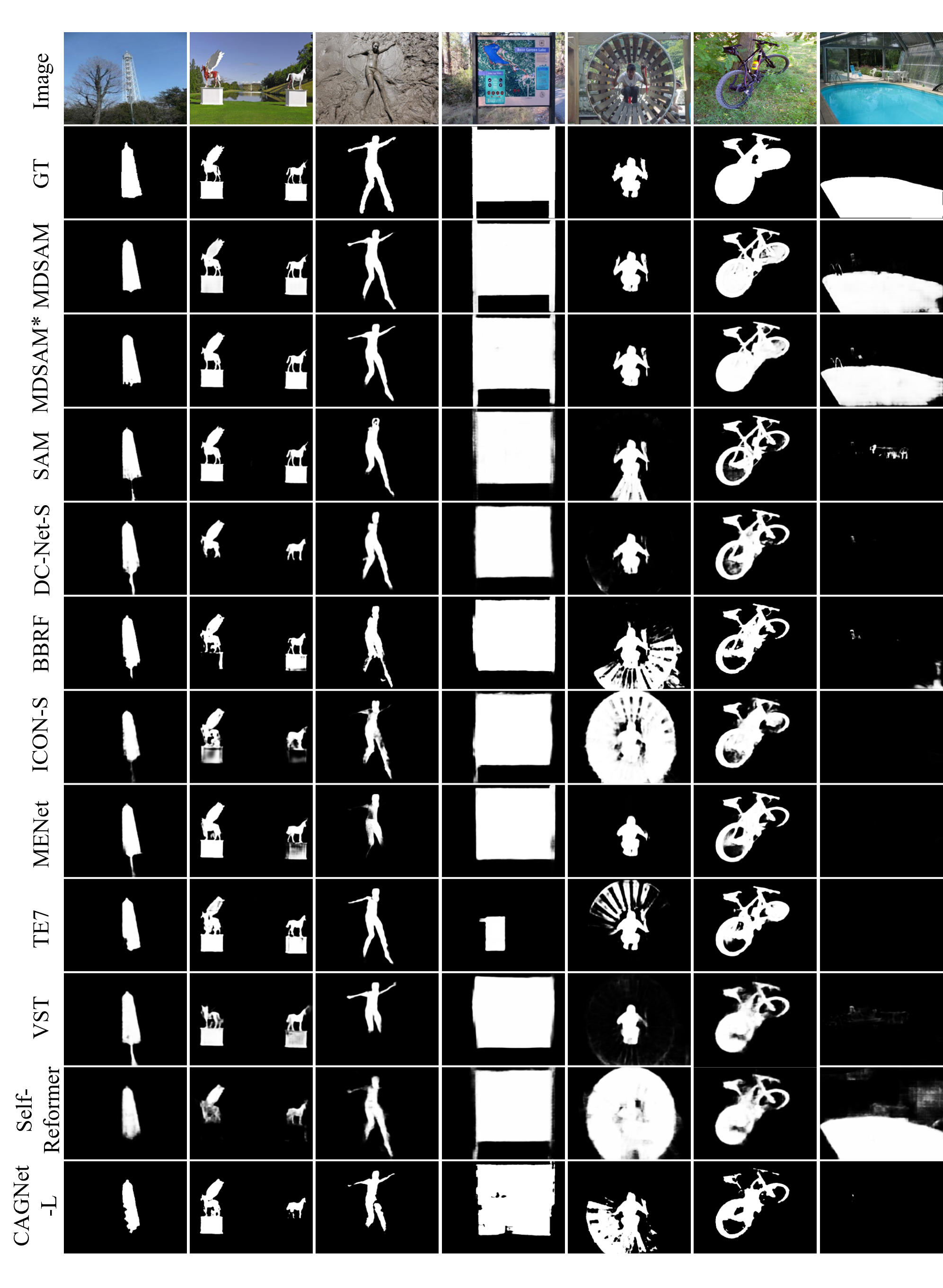}
\caption{Visual comparison of saliency maps with our model and 9 representative methods on five SOD datasets.}
\label{fig:sp3}
\end{figure*}

\begin{figure*}[t!p]
\centering
\includegraphics[scale=0.38]{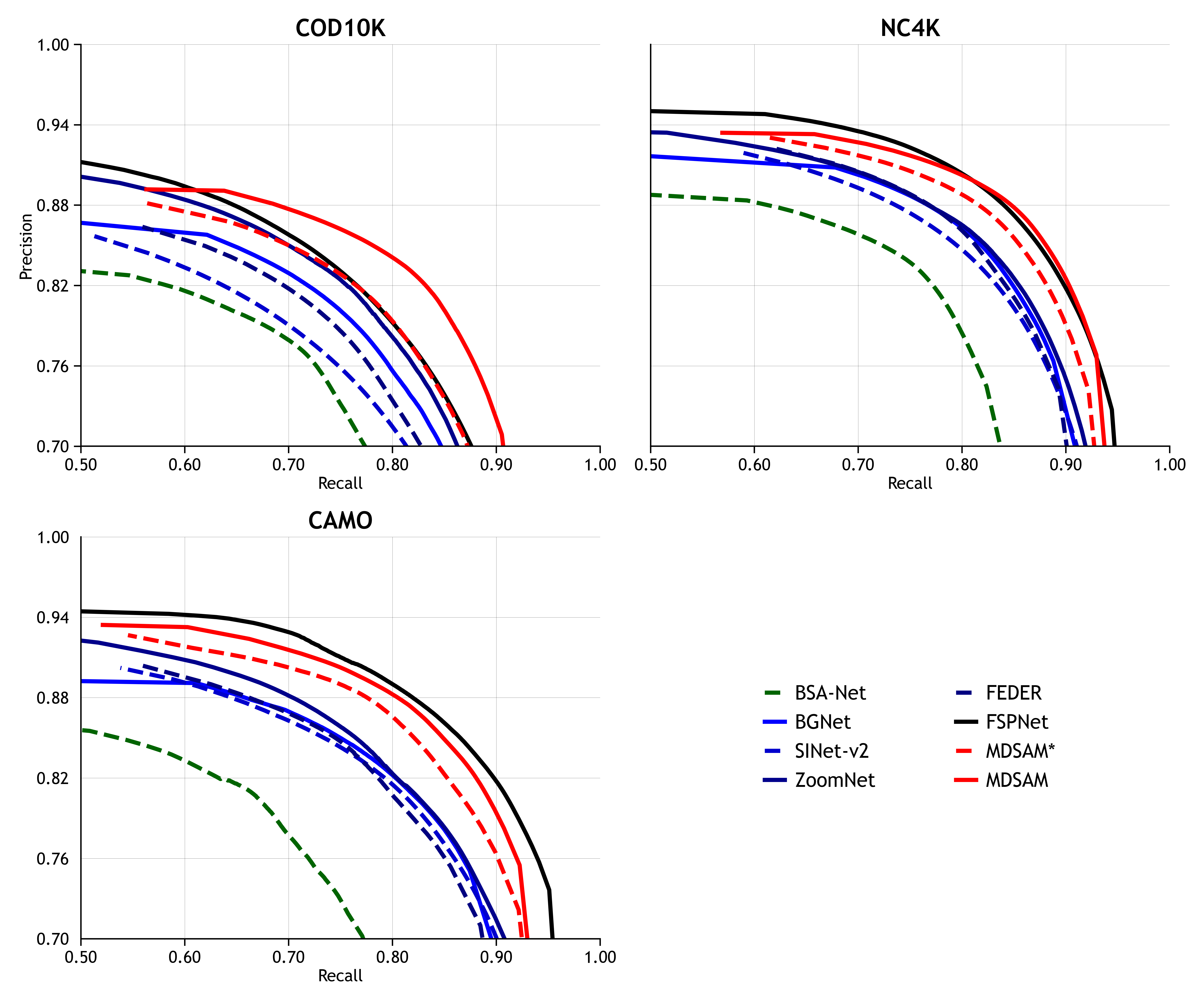}
\caption{Precision-Recall curves comparison on three COD datasets.}
\label{fig:cod_pr}
\end{figure*}

\begin{figure*}[t!p]
\centering
\includegraphics[scale=0.38]{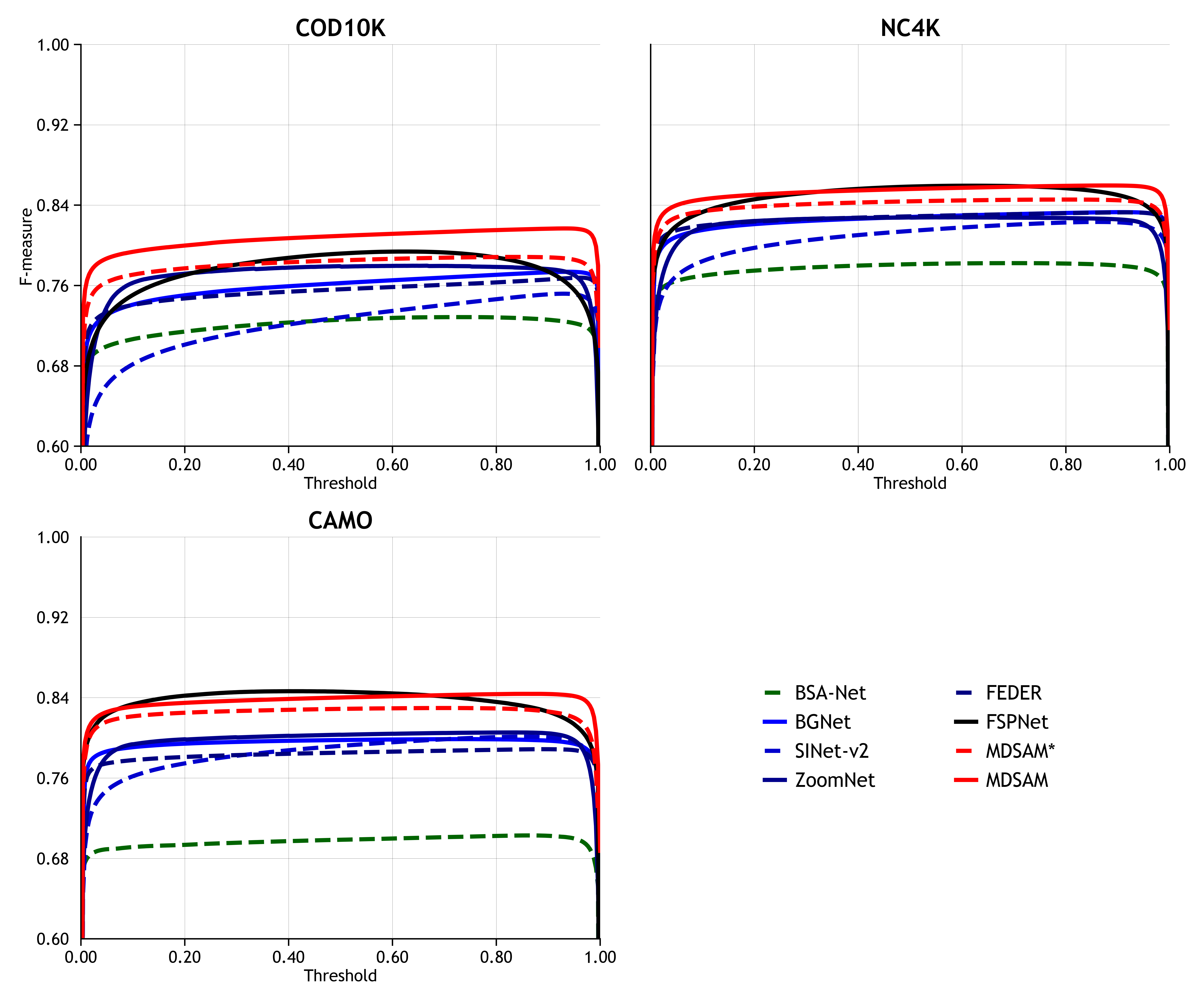}
\caption{F-measure curves comparison on three COD datasets.}
\label{fig:cod_fm}
\end{figure*}

\begin{figure*}[t!p]
\centering
\includegraphics[scale=0.54]{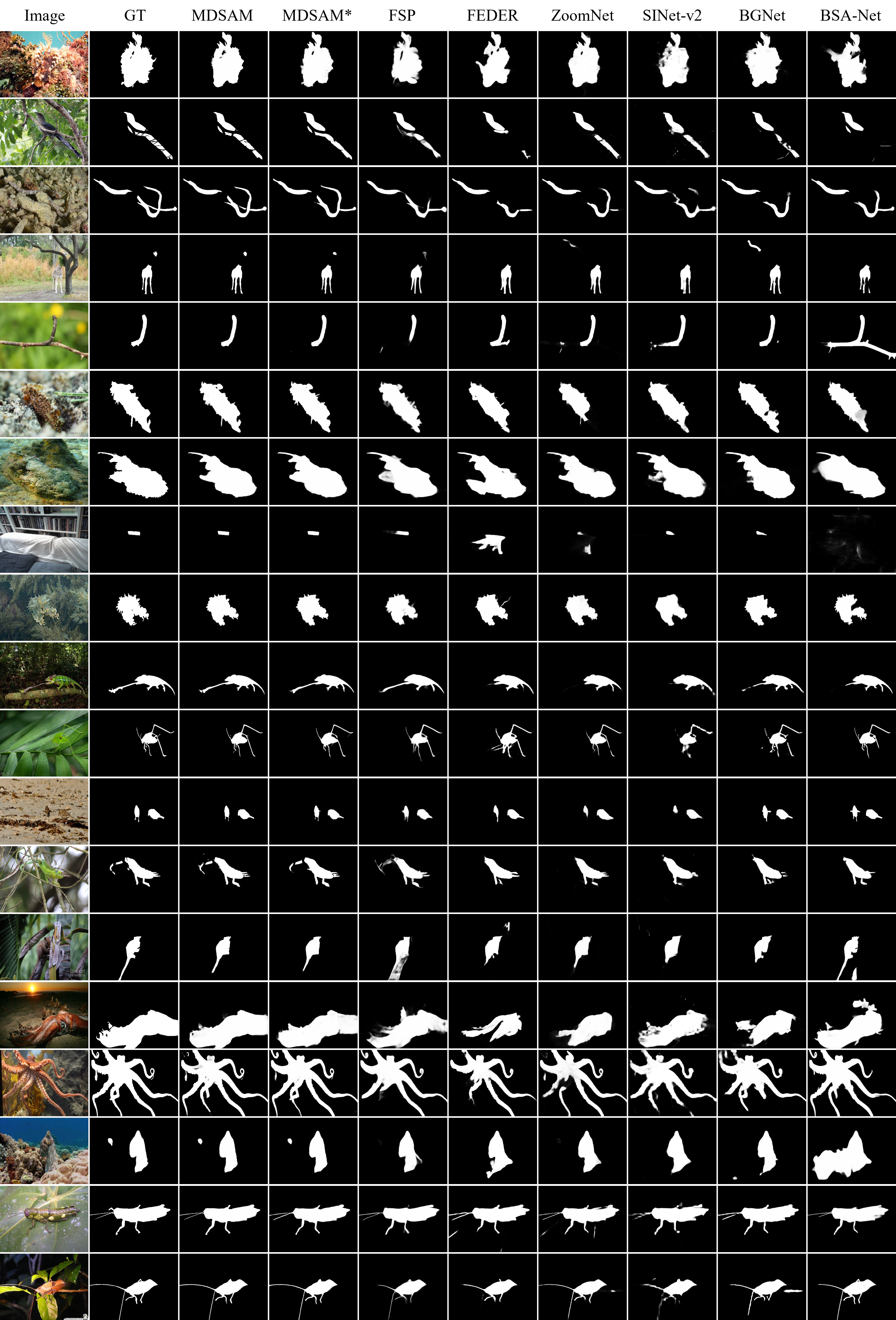}
\caption{Visual comparison of segmentation maps with our model and 6 representative methods on three COD datasets.}
\label{fig:cod_sp1}
\end{figure*}

In the main paper, we compare our MDSAM with other methods by four widely-used metrics. 
In this section, we present the precision-recall curves and F-measure curves in Fig.~\ref{fig:sod_pr} and Fig.~\ref{fig:sod_fm}, compared with CAGNet-L~\cite{mohammadi2020cagnet}, TE7~\cite{lee2022tracer}, MENet~\cite{wang2023pixels}, VST~\cite{liu2021visual}, SelfReformer~\cite{yun2022selfreformer}, ICON-S~\cite{zhuge2022salient}, BBR~\cite{ma2023boosting}, DC-Net-S~\cite{zhu2023dc} and full fine-tuned SAM~\cite{kirillov2023segment} on five SOD datasets.
In addition, we provide more visual comparisons in Fig.~\ref{fig:sp1}, Fig.~\ref{fig:sp2} and Fig.~\ref{fig:sp3}.
In Fig.~\ref{fig:cod_pr} and Fig.~\ref{fig:cod_fm}, we display these curves to compare with SINetv2~\cite{fan2021concealed}, BSA-Net~\cite{zhu2022can}, BGNet~\cite{sun2022boundary}, ZoomNet~\cite{pang2022zoom}, FEDER~\cite{he2023camouflaged} and FSPNet~\cite{huang2023feature} on three COD datasets.
More visual comparisons are shown in Fig.~\ref{fig:cod_sp1}.

\clearpage

\end{document}